\documentclass[a4paper, 10pt, conference]{ieeeconf}
\IEEEoverridecommandlockouts % This command is only needed if you want to use the \thanks command
\usepackage{graphicx}
\usepackage{color}
\usepackage{mathtools}
\usepackage{amsmath} % assumes amsmath package installed
\usepackage{amssymb}  % assumes amsmath package installed
\usepackage{tabularx} % for tables with columns that automatically stretch to the column/page width
\usepackage{booktabs} % to fine-tune tables
\usepackage{multirow} % for multirow cells in a table
\usepackage{subcaption} % for grid-like figure layouts
\usepackage{caption} % to left-align all captions, no matter if single or double row
\usepackage{placeins} % to force figures into section \FloatBarrier
\usepackage{dirtytalk} % set correct quotation marks with command \say
\usepackage[bookmarks=false]{hyperref} % for correctly formatted urls 
\usepackage{tikz}
\usepackage[all, color]{xy}
\usetikzlibrary{positioning,fit,arrows,automata,shapes,positioning,decorations.pathreplacing,matrix}
\usepackage{xcolor}
\definecolor{cPlayerEmpowerment}{HTML}{5e3c99}
\definecolor{cCompanionEmpowerment}{HTML} {FF0000}%{b2abd2}
\definecolor{cCPTransferEmpowerment}{HTML}{FF0000}
\definecolor{cPlayer}{HTML}{5e3c99}
\definecolor{cEnemy}{HTML}{e66101}
\definecolor{cGray}{HTML}{848484}

\newcommand{\Ex}{\mathbb{E}}

\newcommand{\tableFig}[1]{\multirow{2}{*}{\begin{minipage}[b]{0.7cm}\vspace{-0.05cm}\includegraphics[width=0.7cm]{#1}\vspace{0.1cm}\end{minipage}}}

\title{\LARGE\bf
New And Surprising Ways to Be Mean:\\Adversarial NPCs with Coupled Empowerment Minimisation
}

\author{Christian Guckelsberger$^{1,*}$, Christoph Salge$^{2,3}$ and Julian Togelius$^{3}$% <-this % stops a space
\noindent\thanks{$^{1}$Computational Creativity Group, Department of Computing, Goldsmiths, University of London, London, SE14 6NW, UK.}%
\noindent\thanks{$^{2}$Adaptive Systems Research Group, School of Computer Science, University of Hertfordshire, Hatfield, AL10 9EU, UK.}% 
\noindent\thanks{$^{3}$Game Innovation Lab, Department of Computer Science and Engineering, New York University, Brooklyn, 11201, USA.}% 
\noindent\thanks{$^{*}$Corresponding email: {\tt\small c.guckelsberger at gold.ac.uk}}
}

\begin{document}

\maketitle
\thispagestyle{empty}
\pagestyle{empty}

\begin{abstract}
Creating Non-Player Characters (NPCs) that can react robustly to unforeseen player behaviour or novel game content is difficult and time-consuming. This hinders the design of believable characters, and the inclusion of NPCs in games that rely heavily on procedural content generation. We have previously addressed this challenge by means of empowerment, a model of intrinsic motivation, and demonstrated how a coupled empowerment maximisation (CEM) policy can yield generic, companion-like behaviour. In this paper, we extend the CEM framework with a minimisation policy to give rise to adversarial behaviour. We conduct a qualitative, exploratory study in a dungeon-crawler game, demonstrating that CEM can exploit the affordances of different content facets in adaptive adversarial behaviour without modifications to the policy. Changes to the level design, underlying mechanics and our character's actions do not threaten our NPC's robustness, but yield new and surprising ways to be mean. 
\end{abstract}

\section{Introduction}

Non-Player Characters (NPCs) in video games serve many purposes: they can be quest givers, conversation partners, leaders, sidekicks or other kinds of collaborators \cite{warpefelt2016non}. But in many cases they are \emph{adversaries}. Adversarial NPCs also come in many forms, their behaviour varying according to the game genre, the design affordances, and the underlying algorithms. Treanor et al.~\cite{treanor2015ai} make the fundamental distinction between \emph{AI as Adversary} and \emph{AI as Villain}. Adversaries are designed to defeat the player without resorting to cheating, e.g. an AI for Chess or Go. The objective of an NPC villain in contrast is not to defeat the player but to create an interesting challenge which can be overcome eventually. We refer to both types simply as adversaries.

Irrespective of the type, an NPC's primary goal is usually to convey a special player experience. A substantial part of this experience is shaped by the believability of their behaviour \cite{warpefelt2016non}: a believable adversary can, amongst others, adapt to changes in the world and allows the player to attribute goal-ownership. In existing adversary AI however, these attributes are either not present, or very shallow.

NPCs in video games are largely hand-authored, using representations such as finite state machines and behaviour trees. This limits most NPC AI to a particular game and a specific role. While such NPCs might appear to own their goals, they will hardly adapt to unanticipated player behaviour or changes in the game world. The latter aspect is partly alleviated by reinforcement learning, evolutionary approaches or planning. However, there is two caveats. Algorithms such as Monte Carlo Tree Search (MCTS) are typically targeted towards maximising adversarial efficacy against the player, resulting in blunt and single-faceted behaviour. Procedural personas~\cite{holmgaard2018automated} contribute to the impression of more multi-faceted behaviour by optimising a set of pre-specified utilities. However, even these advanced approaches usually rely on objective functions, rewards and training samples which are strongly tied to specific affordances of the game world. As soon as this world changes, the basis for their behaviour and thus their believability is lost.

An alternative approach is to use models of intrinsic motivation \cite{Oudeyer2008} to drive NPC behaviour. Models of intrinsic motivation do not rely on externally specified rewards, and thus allow an agent to act sensibly even if its means to interact with the world change. As intrinsic motivations are usually aligned with key drivers of agency, this can give the appearance of goal-directness. Merrick and Maher~\cite{Merrick2009} have used curiosity and learning progress as intrinsic reward signals in reinforcement learning to drive NPC behaviour. However, their NPCs work in isolation, and interactions with the player would be incidental and likely shallow. 

In this paper, we address the challenge of creating \emph{generic adversarial NPCs}, i.e.~NPCs that can adapt and respond to substantial changes in the game environment, mechanics and a character's abilities, and that exhibit a wide range of new and surprising adversarial behaviours that are not uniquely focused on winning over the player. We prose to use the intrinsic motivation formalism of \emph{Coupled Empowerment Minimisation} (CEM) \cite{guckelsberger2016c}, an action policy based on the information-theoretic quantity \emph{empowerment} \cite{Klyubin2008}. Empowerment quantifies the options available to an agent in terms of availability and visibility. In the stochastic case, it generalises to an agent's potential and perceivable influence on the game world, including other agents such as the player. Empowerment forms the basis of \emph{empowerment maximisation}, an action policy which drives agents towards states where they have a higher influence on their environment. CEM is an extension of this principle to the multi-agent case. The main idea behind CEM is that an agent not only maximises its own- but also maximises or minimises the empowerment of one or more other characters. In previous work \cite{guckelsberger2016c}, we have exploited the \emph{maximisation case} to formalise companion-like behaviour in a very general and flexible way. We expect the policy to yield sensible NPC behaviour in any game where a player's progress towards a goal is accompanied by an increase in options and influence, and thus empowerment. This is the case for most games: consider e.g.~the effect of accumulating resources and building units in strategy games, collecting inventory items in an RPG, or using power-ups or additional and stronger weapons in a shooter. 

In this work, we look at the \emph{minimisation} case to design more believable adversarial NPCs. Our NPCs essentially choose actions which increase their own-, and decrease the player's empowerment. Note that this is different from simply maximising or minimising a utility such as score or health, and promises to give rise to highly adaptive, unexpected and novel adversarial behaviour. We explore CEM to drive adversarial NPC behaviour in different levels of a turn- and tile-based dungeon-crawler game, where an CEM-driven agent is confronted with changes in the environment and its own abilities. A qualitative analysis demonstrates that CEM yields sensible and interesting adversarial behaviour across a range of game modes. Relating to Treanor et al.~\cite{treanor2015ai}, we show how different parametrisations of our policy give rise to different adversary types, from opportunists to super-villains.

\section{Coupled Empowerment Maximisation}
In a nutshell, a CEM-driven agent acts to maximise its \hbox{own-,} while either maximising or minimising another agent's empowerment. We investigate the minimisation case here, and therefore complement previous work \cite{guckelsberger2016c}. CEM relies on two types of empowerment: \emph{(vanilla) empowerment} as briefly mentioned in the introduction, and the distinct \emph{transfer-empowerment}. We now introduce both quantities formally. Our focus is on games that are discrete in time and space. However, continuous empowerment implementations exist. An extensive survey of motivations, intuitions and past research on empowerment can be found in \cite{Salge2014}. CEM has previously been covered in \cite{guckelsberger2016, guckelsberger2016c} and \cite{salge2017empowerment}. 

\subsection{Empowerment and Transfer Empowerment}
Empowerment is an information-theoretic quantity. It is zero when the agent has no control over what it can perceive, i.e. when all actions lead to the same or a random sensor state, and it increases when different actions lead to separate perceivable outcomes. We represent an agent's actions, its future sensor states, and the state of the environment as random variables $A$, $S$, and $R$, respectively. The causal conditional probability distribution $p(S_{t+1}|A_t, r_t)$ then models the impact of the agent's actions, performed in a specific environment state $R_t=r_t$, on its future sensor states. For the calculation of empowerment, this distribution is interpreted as a memoryless, potentially noisy information-theoretic communication channel. 

\emph{Vanilla empowerment} $\mathfrak{E}_{r_t}$ in a given environment state $r_t$ is calculated as the \emph{channel capacity}, corresponding to the maximum potential information flow that an agent could induce into its future sensor state by a suitable choice of actions. More generally, we consider a sequence of actions $A^n_t=(A_{t}, \dots ,A_{t+n-1})$ corresponding to a lookahead of $n$. With \emph{n-step empowerment} we then measure an agent's influence on its future sensor state $n$ steps in the future:
\begin{equation}
	\mathfrak{E}^n_{r_t} = \max\limits_{p(a^n_t)} I(S_{t+n};A^n_t|r_t)
\end{equation}
The term being maximised represents the mutual information between the actuator and future sensor states, given the current environment state $r_t$. 

\emph{Transfer empowerment} $\mathfrak{E}^{T,n}_{r_t}$ relates the actions and sensor of \emph{two} agents: it quantifies the potential influence the active agent's actions have on the other's future sensor state. The channel capacity underlying both empowerment types can be exactly determined using the Blahut-Arimoto algorithm \cite{Blahut1972,Arimoto1972}. For an introduction to the information-theoretic notions see \cite{Cover2006}, and \cite{Salge2014} for a detailed introduction to empowerment. 

\subsection{Coupled Empowerment Maximisation}
CEM is an extension of empowerment maximisation to the multi-agent case, and we consequently have to account for the actions of other agents. In this paper, we focus on the turn-wise interaction of \emph{one} NPC with the player. Each interaction cycle is initiated by the player performing an action, which the NPC reacts to in the next time step. Both agents can affect the other either explicitly, or implicitly through their impact on the shared game world. We hypothesise that decreasing the player's empowerment gives rise to adversarial behaviour. To test this hypothesis, we model the NPC's policy such that it not only maximises the NPC's own, but also minimises the player's empowerment: 
\begin{equation}
\resizebox{.91\hsize}{!}{$
	\pi(r_t) = \underset{a_t} {\operatorname{arg\,max}}\ \left(\alpha_A \cdot \Ex[\mathfrak{E}^{A,n}_{r_{t+2}}]_{a_t} + \alpha_P \cdot \Ex[\mathfrak{E}^{P,n}_{r_{t+1}}]_{a_{t}} + \alpha_T \cdot \Ex[\mathfrak{E}^{T,n}_{r_{t+2}}]_{a_{t}} \right)
	$}
\end{equation}
Here, parameters $\alpha_A, \alpha_P$ and $\alpha_T$ weight the expected adversary, player and transfer $n$-step empowerment in the overall coupling. The \emph{adversary-player transfer empowerment} serves the maintenance of \emph{operational proximity}: even if the NPC cannot affect the player's empowerment at the current point in time, it will try to remain in states where it can at least affect the player's perception, and thus increase the likelihood of affecting its empowerment in the future. 

Determining the optimal action is a two-stage process. For the first stage, we have to note that empowerment is a state dependent quantity and the policy thus involves expectations over the NPC's actions. For the computation of these policies, the NPC first determines which environment states its own actions could yield at $t+1$. This is where the player acts next, and where player empowerment will be computed. For the calculation of its own and adversary-player transfer empowerment however, it needs to anticipate the consequences of the player's actions on the distribution of environment states at $t+2$. In the second stage, the NPC then calculates player empowerment in $t+1$, as well as its own and transfer empowerment in $t+2$. This requires another $2n$ rounds of anticipation steps. For the calculation of empowerment, the resulting environment states are transformed to potentially limited sensor states $S$. 

Unlike algorithms such as Minimax or MCTS, states are uniformly expanded up to a fixed depth, and only distinguished in terms of whether they are perceived differently. Here, we model the player's policy as a uniform distribution.

\subsection{Health-Performance Consistency}
In many games, a decrease in health or a different core game quantity does not necessarily result in the decline of a character's abilities. For short lookaheads $n$, empowerment would thus remain unaffected. We counteract this by adopting the transformation of \emph{health-performance consistency} (HPC) from previous work \cite{guckelsberger2016c}. It reduces the probability of a character's action to lead into the follow-up state originally prescribed by the environment dynamics proportionally to its remaining health. The more an agent's health decreases, the more likely its actions will be ineffective. Formally:
\begin{equation*}
p(r_{t+1}|a_t, r_t) =
\begin{cases}
1-\gamma + \gamma p(r_{t+1}|a_t, r_t) , & \text{if } r_{t+1} = r_t \\
\gamma p(r_{t+1}|a_t, r_t) & \text{else.}
\end{cases}
\end{equation*}
Here, $\gamma = h_t / h_{\textit{max}}$, the ratio of the agent's remaining and maximum health. This can of course be modified to model non-linear changes. We use HPC as a means for optimisation, but it is not a necessity: given a large enough lookahead $n$, the long-term effect of decreasing an agent's health will be reflected in its empowerment.

\begin{table*}[ht!]
	\small
	\caption{Level elements in dungeon-crawler testbed}
	\vspace{-0.2cm}	
	\label{tbl:levelElements}
	\centering
	\renewcommand{\arraystretch}{1.8}
	\begin{tabularx}{\textwidth}[t]{l l p{7cm} X} 			
		\toprule
		\textbf{Sprite} & \textbf{Type} & \textbf{Dynamics} & \textbf{Reason for inclusion}\\ \hline
		\tableFig{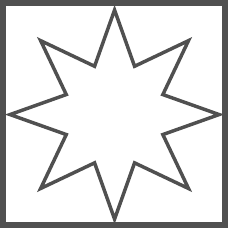} & Goal & Once the character moves on this tile, the game is won. \hphantom{additional spacing to expand to two lines} & To provide a gradient for progression within a level.\\ \hline
		\tableFig{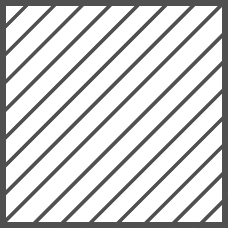} & Wall & Immovable obstacle. Cannot be penetrated by attacks, and hides anything behind from character's perception. & To structure level and provide choke points for specific interactions. Allows for discovery of hidden elements.\\ \hline
		\tableFig{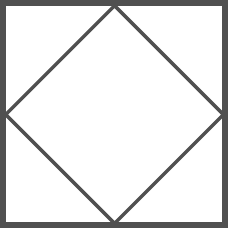} & Lava & Decreases a character's health by a fixed amount for each time step it remains on the field. & Structures environment further and introduces health trade-offs. Allows for rich interaction with pushing.\\ \hline
		\tableFig{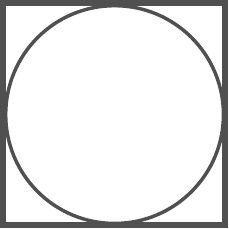} & Recharger & Increases a character's health by a fixed amount for each time step it remains on the field. & Makes health a manageable and expendable resource that might be traded for other gains.\\ \hline
		\tableFig{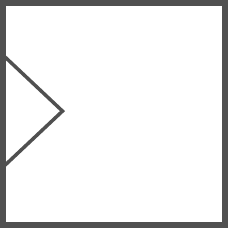} & Turret & Shoots arrow, inflicting a fixed health damage on the first character being hit. Here pointing east. & Serves as threat separate from characters' attack facilities, and bears danger of self-inflicted damage. \\ \hline
		\tableFig{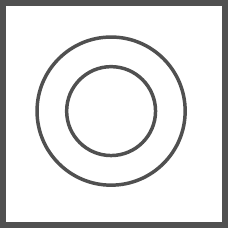} & Trigger & Activates connected turret for each time step that a character remains on the tile. & Triggers can be far off the activated turret and thus allow to strike remotely.\\ 
		\bottomrule
	\end{tabularx}
\end{table*}

\section{Evaluation}
We hypothesise that CEM yields highly adaptive, adversarial behaviour. The behavioural dynamics following changes to the game might be surprising even for the game's designers, and likely increase the believability of our characters. However, quantitative means to evaluate gameplay and player experience provide insufficient evidence, as they cannot capture these dynamics as well as novelty and believability in sufficient detail. We consequently perform a qualitative, exploratory study, providing the necessary insights for a quantitative study to follow in future work. This evaluation is complemented with online videos of the NPC behaviour. 

\begin{table*}[ht!]
	\small
	\caption{Character abilities in dungeon-crawler testbed}
	\vspace{-0.2cm}	
	\label{tbl:characterAbilities}
	\centering
	\renewcommand{\arraystretch}{1.3}
	\begin{tabularx}{\textwidth}[t]{l p{7cm} X} 			
		\toprule
		\textbf{Action} & \textbf{Dynamics} & \textbf{Reason for inclusion}\\ \hline
		Idle & Causes no change to the current game state. & Represents fallback if other actions are disadvantageous. \\ \hline
		Move & Move non-diagonally into adjacent cell if there is no obstacle. Otherwise only changes character orientation. & Common mechanic allowing for exploration, hiding and change of position as reaction to other characters.\\ \hline
		Push & In addition to moving, shift adjacent characters in the movement direction if there is no obstruction. & Allows for complex interactions with the environment by pushing others into lava, rechargers, or a turret's target range.\\ \hline
		Fly & Allows to move over lava fields without taking damage. The character can still benefit from rechargers. & A way to access previously inaccessible parts of a level, and make other characters face new obstacles.\\ \hline
		Melee attack & Causes damage to adjacent characters if being faced. The amount of health damage is predefined. & Common mechanic for predator-and-prey scenarios. Requires to run away or attack from a distance before others close. \\ \hline
		Range attack & Reduces health of first character in current direction within attack range. Damage and range are predefined. & Allows to imbalance attack options based on spatial proximity, making seeking cover a sensible move to escape damage. \\ \hline
		Heal & Increases health of adjacent, faced character by fixed amount and up to maximum health for that character. & To check if an action which conventionally does not feature in player-adversary interaction is chosen and to what effect.\\
		\bottomrule
	\end{tabularx}
\end{table*}

\subsection{Method}
We conduct three individual experiments to test our hypothesis and investigate how a CEM-driven NPC copes with increasingly tough challenges in game development and research: (1) predator-and-prey behaviour as present in many games, (2) the exploitation of affordances in the agent-environment interaction, and (3) the ability to interact with the player from a distance. The latter is controlled by the experimenters. To probe how CEM contributes to novelty and adaptivity, we change the environment dynamics and the abilities of both characters and analyse the emergent behaviour. Each experiment comprises a number of scenarios.

Due to the richness of our testbed, an exhaustive search through the space of environmental features and character abilities would be infeasible. We consequently focus on those combinations that yield the biggest difference in emergent behaviour. The CEM parameters, i.e. the weights $\alpha$ and the agent's lookahead $n$, cannot be evaluated exhaustively either in a qualitative study. Instead, we highlight how specific configurations allows us to model different adversary types, thus stressing the opportunities in parameter fine-tuning.

\subsection{Testbed}
We have adopted our dungeon-crawler testbed from previous work on CEM-driven general companion NPCs \cite{guckelsberger2016c} with the goal to support comparisons and provide a basis for a future joint quantitative evaluation. The testbed is discrete in time and action/state space, which simplifies the computation of our policy and the analysis of behaviour.

The game is populated by the player character and one CEM-driven NPC. Characters interact in turn-wise order, and the player must navigate to a goal-tile to win the game. All characters have a current and maximum amount of health points, which are indicated by numbers at the bottom of their avatars. To provide rich challenges for adaptation, we have extended our previous testbed substantially with both new environmental features and character abilities. Tbl. \ref{tbl:levelElements} and \ref{tbl:characterAbilities} provide an overview of the various features/abilities, their dynamics and the rationale behind their inclusion.

The sensors of player and adversary are \emph{asymmetric}, \emph{local} and \emph{non-overlapping}. They are asymmetric in that the player can also perceive the game status, while the NPC cannot. Locality means that e.g.~the NPC only perceives the player or other dynamic game elements within a fixed radius. This determines a character's perceptive field, which is only constrained by walls. Other characters within that field are sensed by their id and relative position. In addition to its own position, sensors only comprise the agent's own rotation and health, but do not allow introspection into other characters. This separation is crucial to avoid overlap between empowerment types. 

We assume a \emph{default configuration} of our agents which is adapted in the experiments. In this minimal setup, characters can only idle and move. Their sensor is limited to a three-cell radius. Furthermore, they are initialised with two of two health points ($h_t, h_{\textit{max}} = 2$). This allows them to take damage without dying right away, and to make use of rechargers. We compute empowerment for a 3-step lookahead ($n{=}3$), and assume an initial weighting of $\alpha_A{=}.5$, $\alpha_P{=}{-}.5$ and $\alpha_T{=}.1$. In this initial setup, the CEM-driven NPC bases its decision-making on the maximisation of its own, and on the minimisation of the player's empowerment to the same extent. We later deviate from this equilibrium and show how unbalanced configurations yield radically different behaviours. Parameters are then chosen from $\alpha_A,\alpha_T \in [0,1]$ and $\alpha_P \in [-1,0]$. In our experiments, actions are always chosen greedily with respect to the policy. We only report on these settings if they deviate from the default configuration. 

\subsection{Experiment 1: Predator-and-Prey}
\begin{figure*}[htbp]
	\centering
	\resizebox{1.003\linewidth}{!}{
	\begin{subfigure}[b]{.24\linewidth}
		\includegraphics[width=\linewidth]{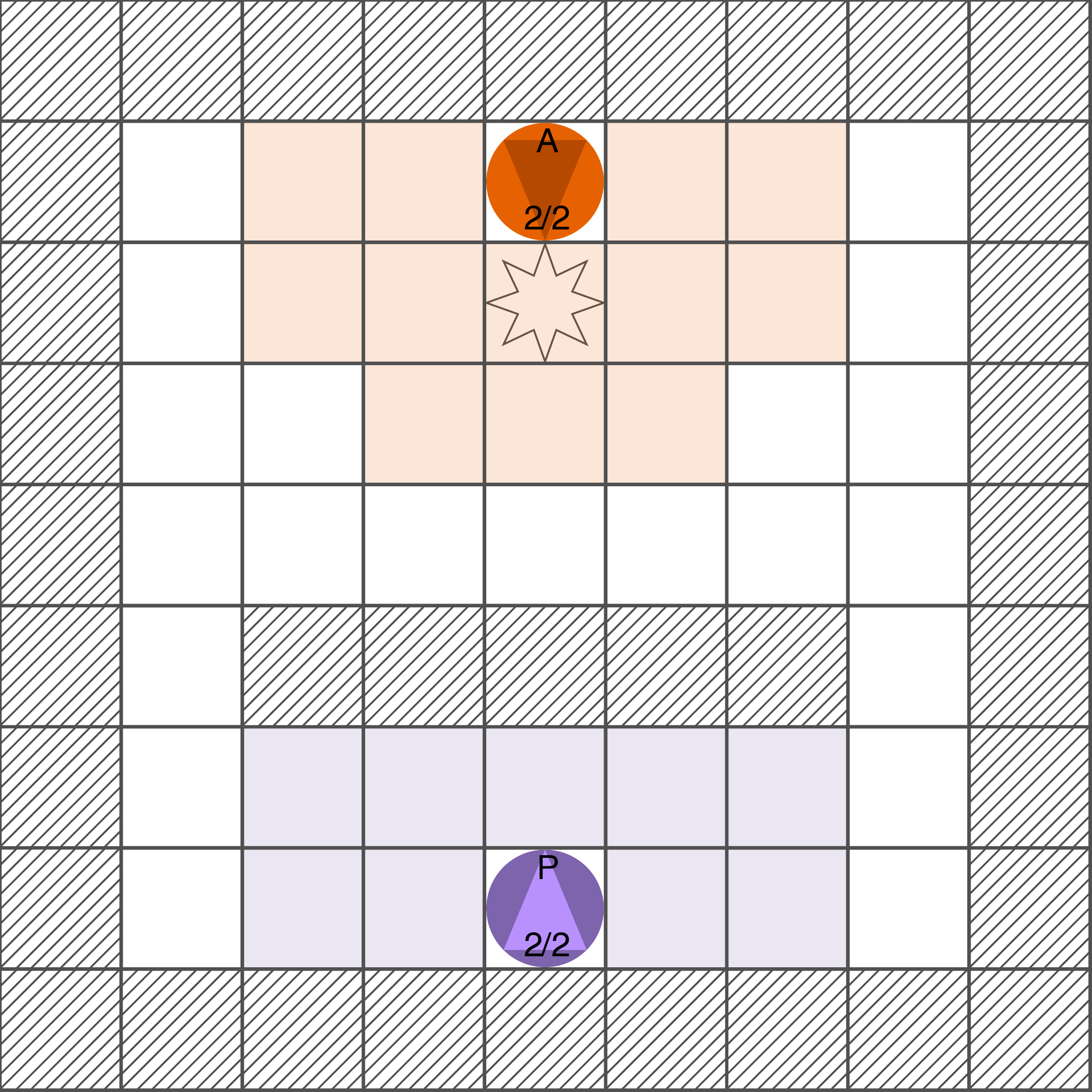}\vspace*{0.03cm}
		\caption{Initial state}\label{fig:experiment_cover_initial}
	\end{subfigure}
	\begin{subfigure}[b]{.24\linewidth}
		\includegraphics[width=\linewidth]{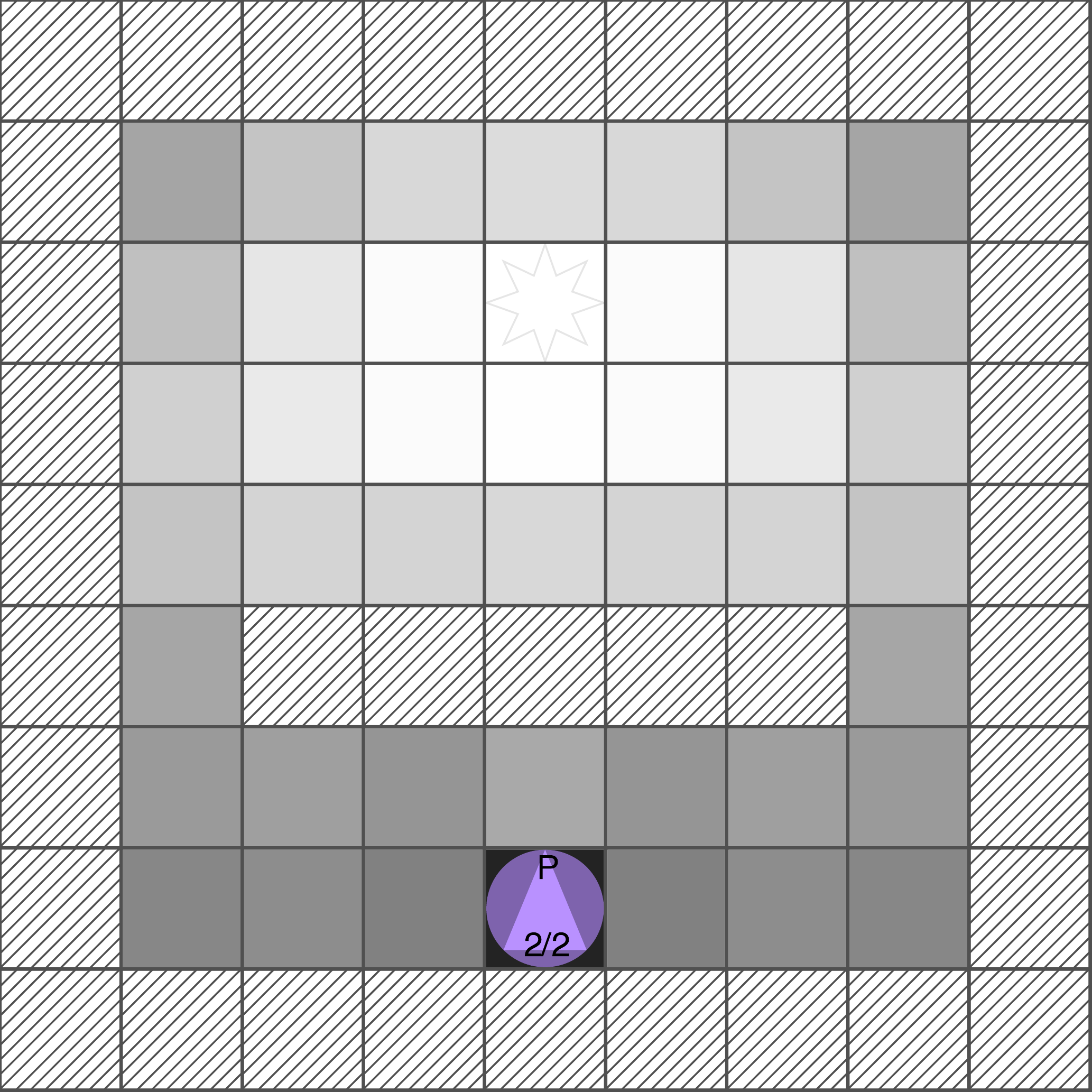}
		\caption{$\mathfrak{E}^{A,3}$}\label{fig:experiment_cover_eE_n3}
	\end{subfigure}
	\begin{subfigure}[b]{.24\linewidth}
		\includegraphics[width=\linewidth]{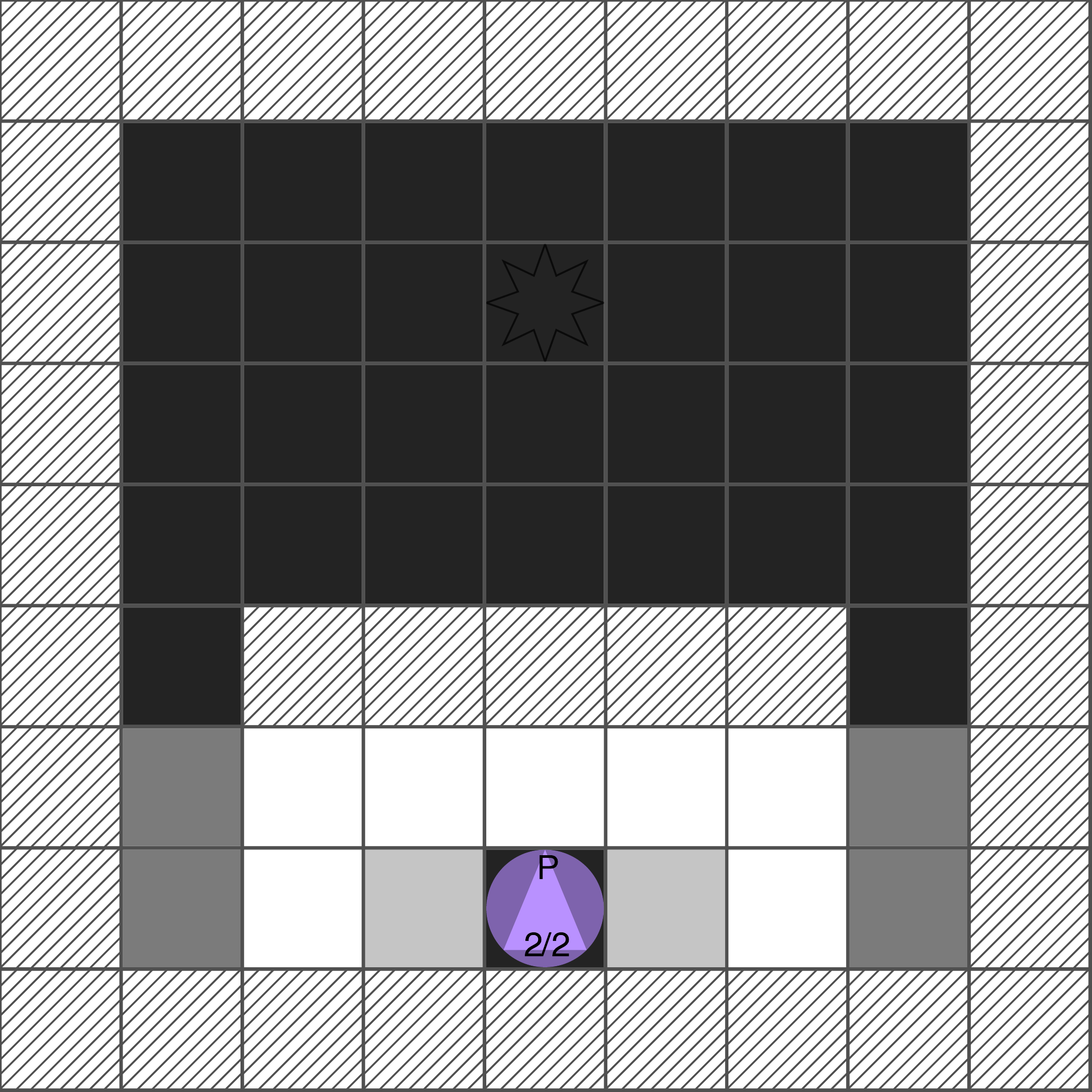}
		\caption{$\mathfrak{E}^{T,1}$}\label{fig:experiment_cover_eT_n1}
	\end{subfigure}
	\begin{subfigure}[b]{.24\linewidth}
		\includegraphics[width=\linewidth]{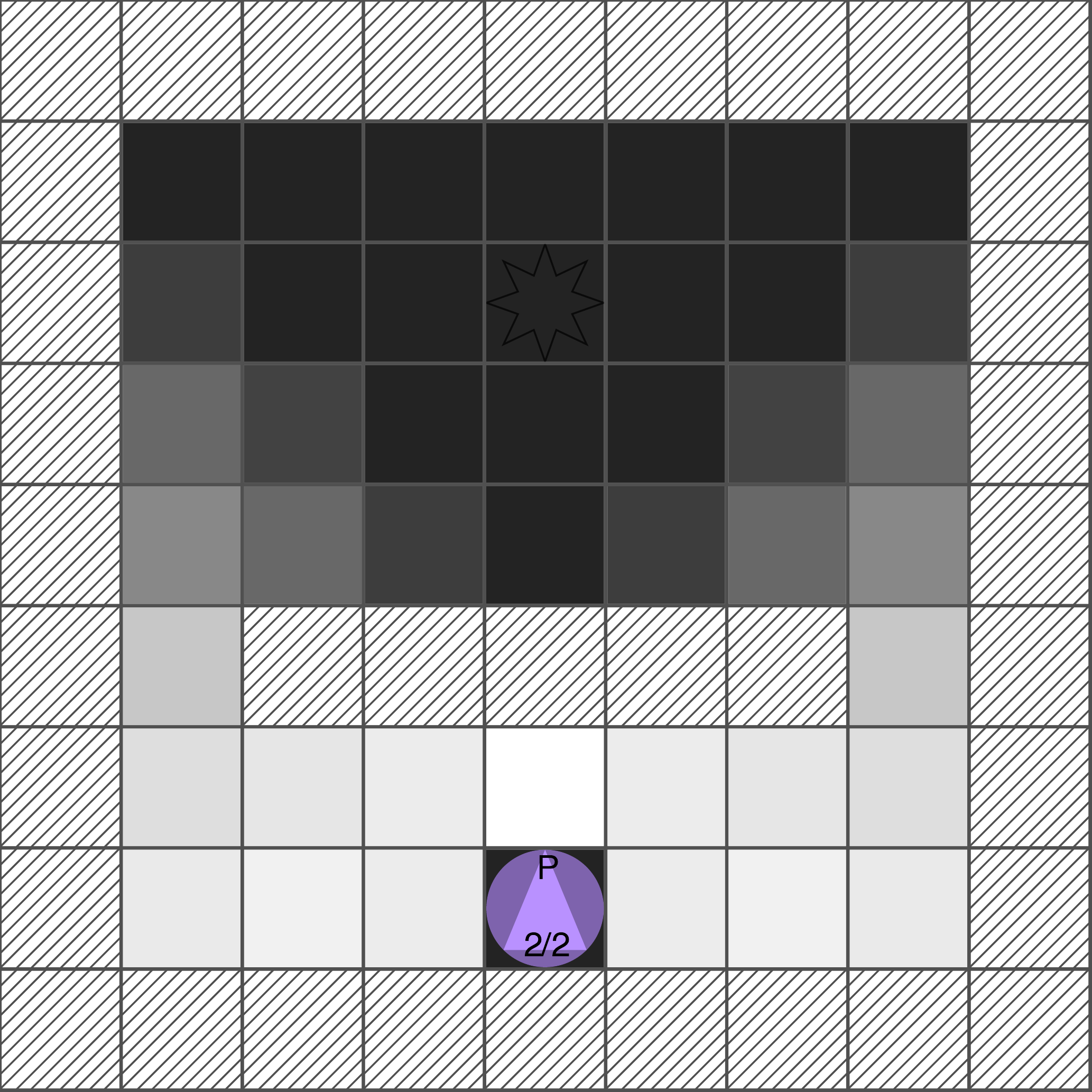}
		\caption{$\mathfrak{E}^{T,3}$}\label{fig:experiment_cover_eT_n3}
	\end{subfigure}
	}
	\caption{Experiment 1. Initial state with perceptive field of adversary and player, followed by adversary ($\mathfrak{E}^{A,n}$) and adversary-player transfer empowerment ($\mathfrak{E}^{T,n}$), the latter for lookaheads $n=1,3$. Brighter hues indicate higher empowerment.}
	\label{fig:exp_cover_maps}
\end{figure*}

The goal of the first experiment is to illustrate the different forces within CEM, and to highlight the policy's potential to give rise to the classic predator-and-prey behaviour which is quintessential to many games. Fig. \ref{fig:experiment_cover_initial} shows the initial state of the environment, consisting of an arena surrounded by walls, divided by a wall with small spaces on the sides to pass through. The adversary ('A', orange) is at the top and faces south, while the player ('P', purple) is situated at the bottom and faces north. Their perceptive field is shown in orange and purple, respectively. 

We have made this environment deliberately simple to familiarise the reader with the different empowerment types in the CEM policy. Fig. \ref{fig:experiment_cover_eE_n3} shows the adversary NPC's empowerment for a 3-step lookahead. Each hue indicates the agent's empowerment if it was moved to that position, but the player's position remained the same. Brighter hues represent higher empowerment. In the default configuration, the agents can only move or idle, and empowerment is consequently very sensitive to degrees of freedom in movement: it is lower where the agent would be blocked, e.g. close to walls and corners. The choke point between the middle and side walls has particularly low empowerment, separating the lower and upper parts of the environment into distinct gradients with local maxima. The player's 3-step empowerment is very similar, given that both agents by default possess the same abilities. Fig. \ref{fig:experiment_cover_eT_n1} illustrates the transfer empowerment from the adversary to the player, for different positions of the adversary. Recall that this empowerment type corresponds to the influence the NPC has on the player's sensor. Hence, for $n{=}1$, it is only non-zero within the player's perceptive field. For larger lookaheads in contrast, it fades out to states from which the NPC could influence the player's perception with some $n$-step action sequences (Fig. \ref{fig:experiment_cover_eT_n3}). This demonstrates that transfer empowerment does not measure perceptibility, but operational-, or in this case, spatial proximity.

\begin{figure*}[htbp]
	\centering
	\resizebox{1.003\linewidth}{!}{\hspace{-0.08cm}
	\begin{subfigure}[b]{.19\linewidth}
		\includegraphics[width=\linewidth]{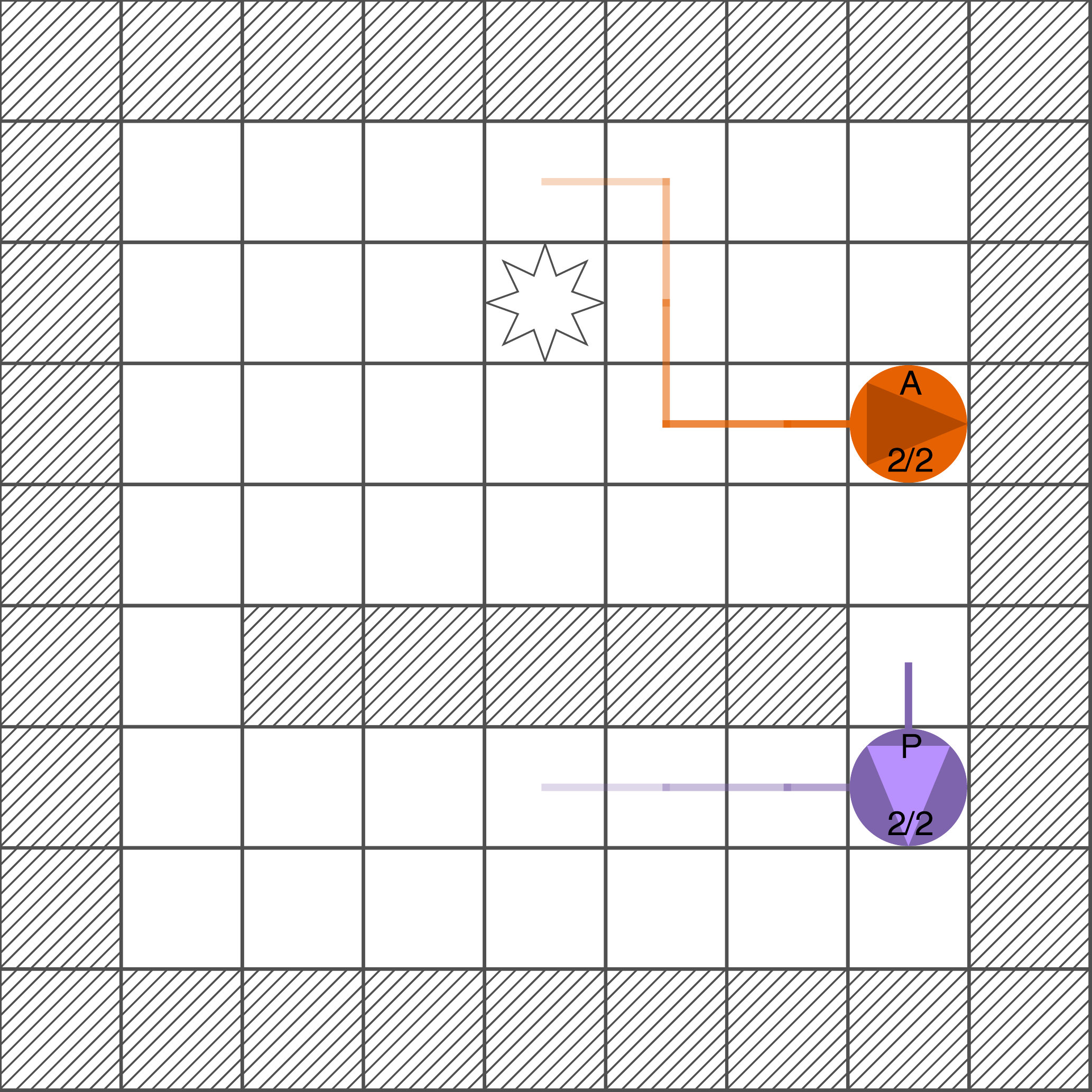}
	\end{subfigure}
	\begin{subfigure}[b]{.19\linewidth}
		\includegraphics[width=\linewidth]{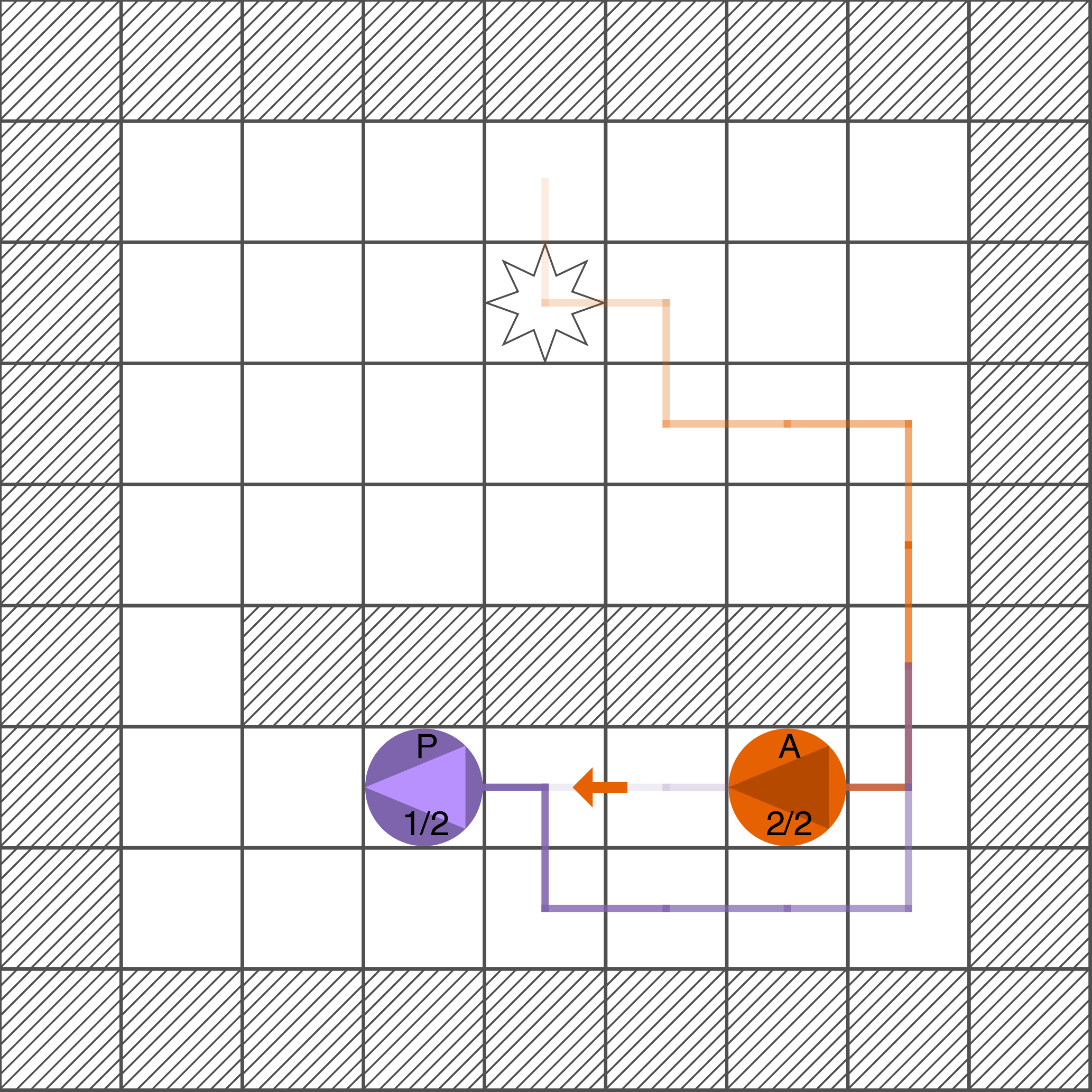}
	\end{subfigure}
	\begin{subfigure}[b]{.19\linewidth}
		\includegraphics[width=\linewidth]{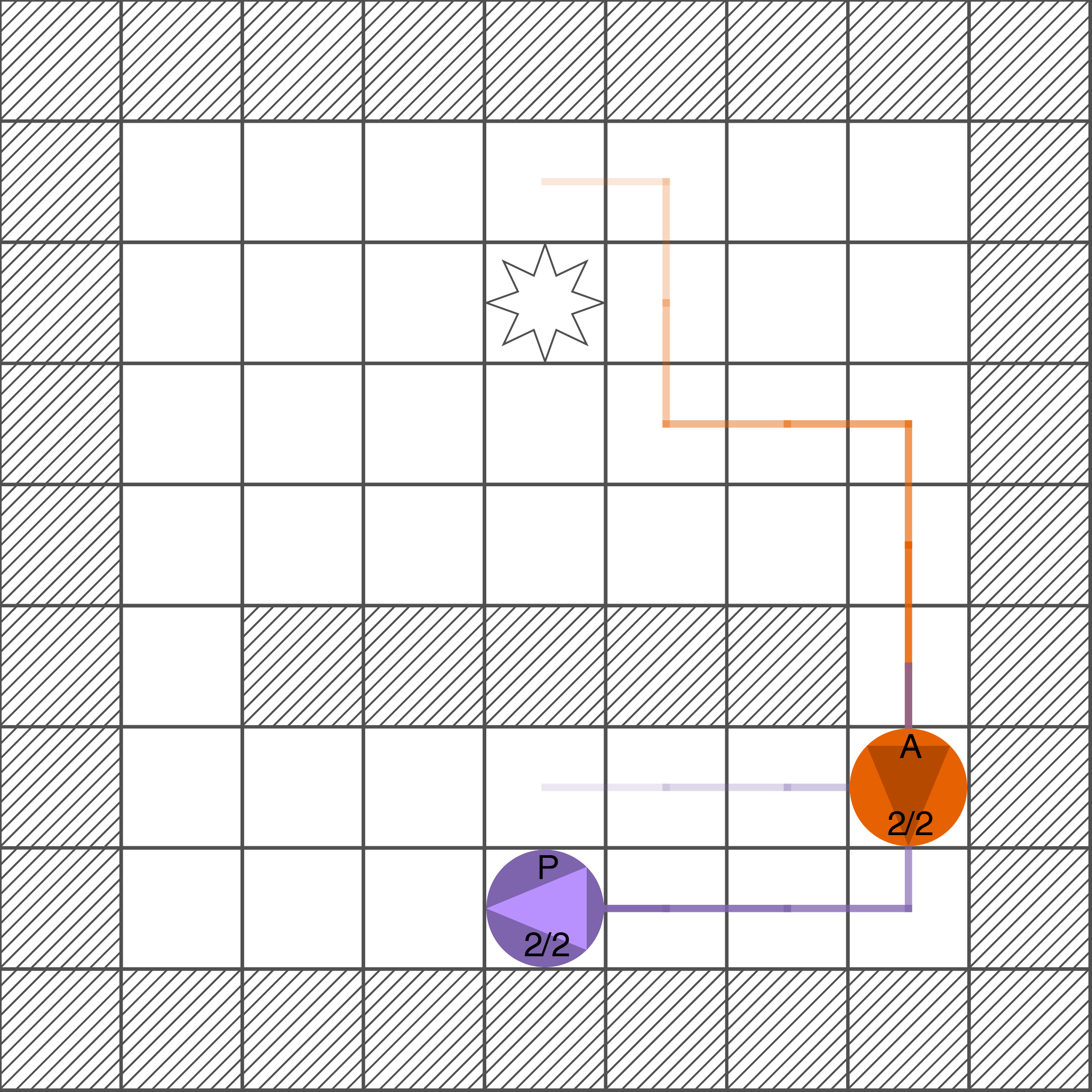}
	\end{subfigure}
	\begin{subfigure}[b]{.19\linewidth}
		\includegraphics[width=\linewidth]{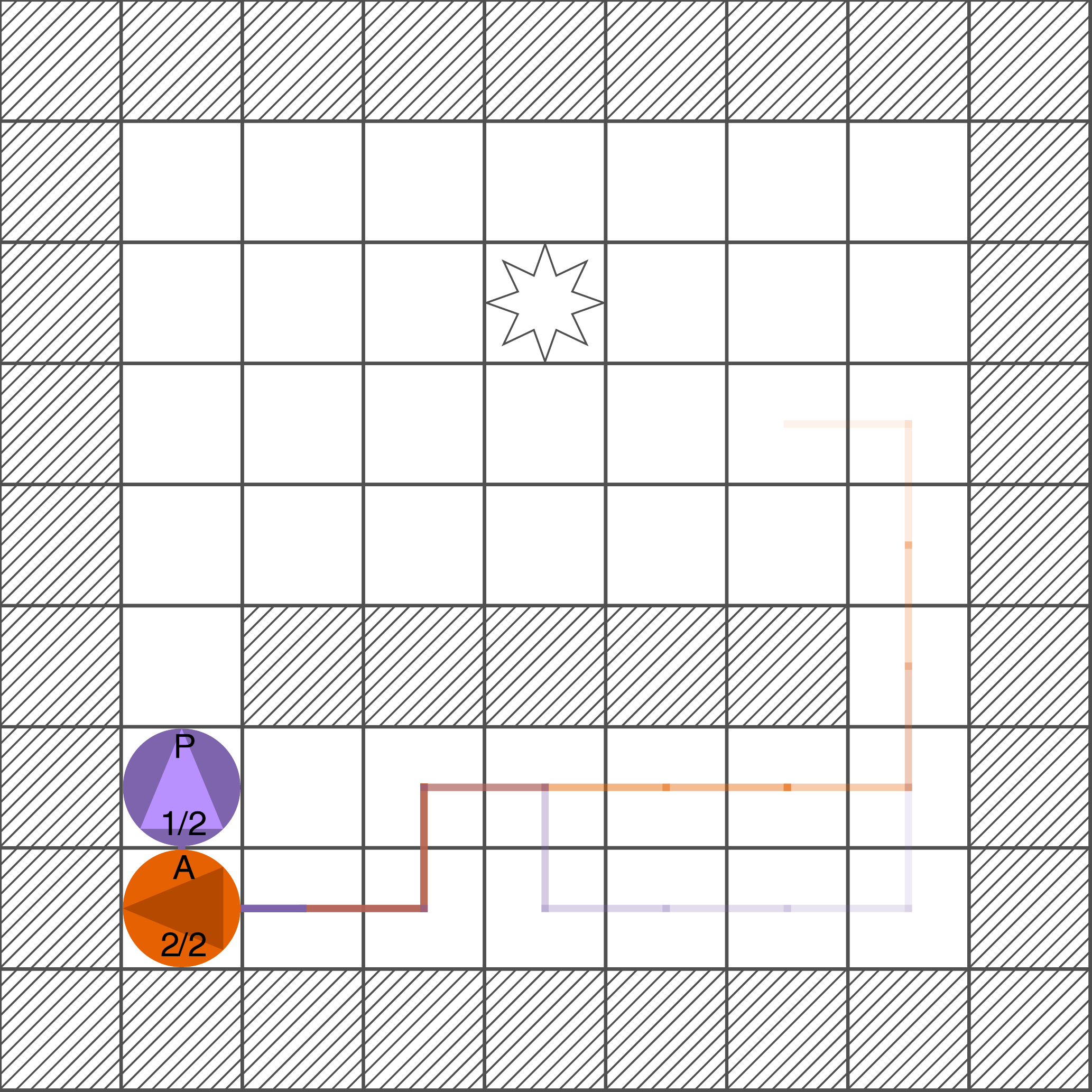}
	\end{subfigure}
	\begin{subfigure}[b]{.19\linewidth}
		\includegraphics[width=\linewidth]{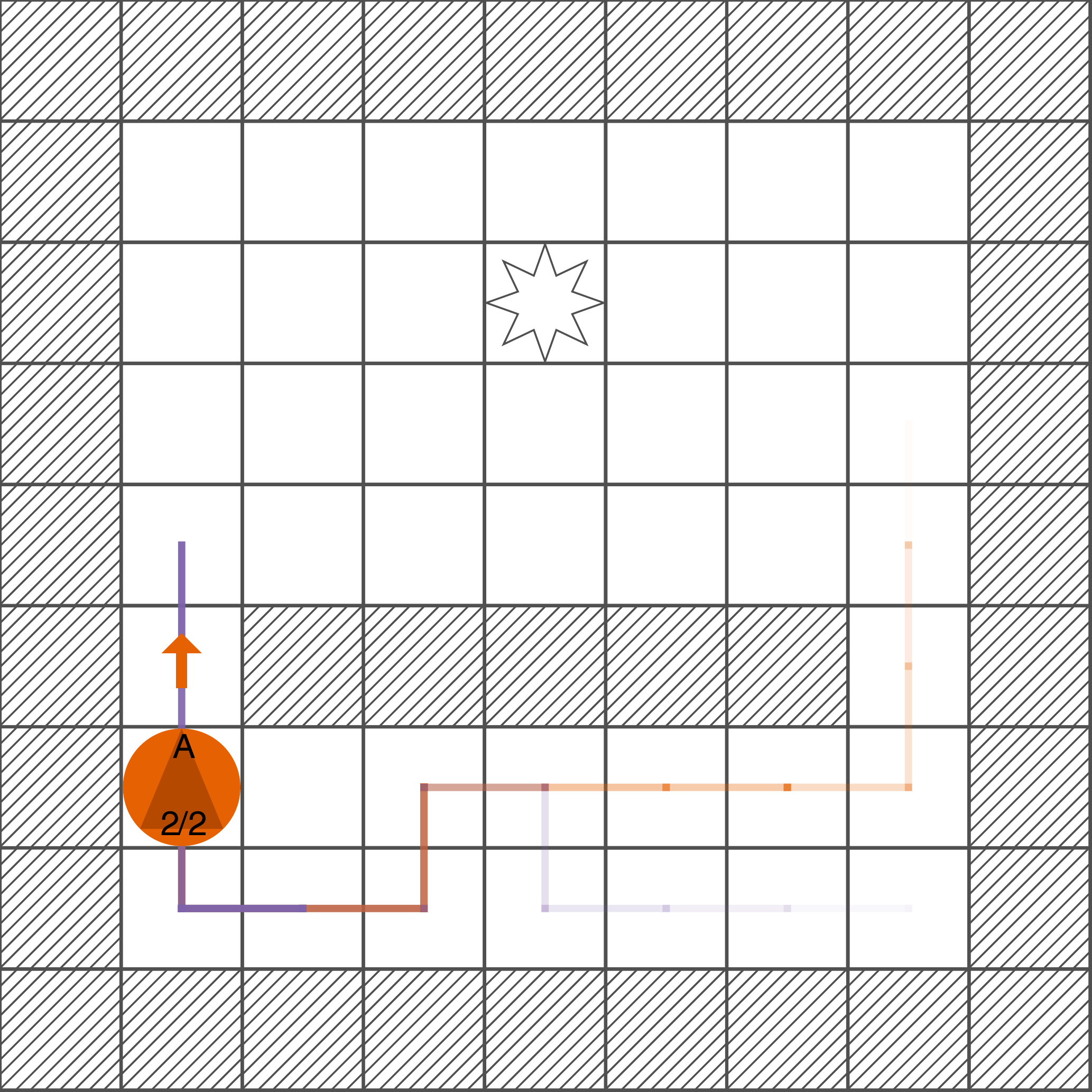}
	\end{subfigure}
	}
	\caption{Experiment 1. ``Daredevil'' adversary ($\alpha_A=.0, \alpha_P=-1.0$ and $\alpha_T=.1$) chasing the player with a range attack.}
	\label{fig:exp_cover_chase_and_kill}
	\vspace{-0.4cm}
\end{figure*}

The contrast between adversary- and adversary-player transfer empowerment highlights how the different empowerment types compete in the CEM policy: If the adversary NPC only considered transfer empowerment, it would move closer to the player; maximising its own empowerment however would require to stay in the middle of the upper part and avoid the choke points on the sides. This trade-off is mediated by the $\alpha$-weights, which can be used to design for different behaviours. Consider the following example: If we equip our adversary NPC with the ability to perform range attacks but stick to the default parameter setup, it remains in the upper area. Nonetheless, if the player moves into this territory, it is killed with two directed shots. We classify this type of adversary as \emph{opportunist}. In contrast, if we increase the negative weight of the player's- while decreasing the weight of the NPC's own empowerment ($\alpha_A{=}.1, \alpha_P{=}{-}1.0$), the NPC is more inclined to trade-off losses in its own- for the decimation of player empowerment. As a result, we get a \emph{daredevil} adversary\footnote{Video online: \url{youtu.be/MVthwbhUNTA}}, chasing and shooting the player as illustrated in Fig. \ref{fig:exp_cover_chase_and_kill}. As final scenario, we investigate the adaptivity of our NPC by equipping the player with a range attack action as well. A video\footnote{Video online: \url{youtu.be/9WoMKJAwl6k}} shows how the adversary adapts to this new threat by dodging and keeping distance. 

This experiment shows that CEM can yield adaptive adversarial behaviour, including the classic predator-and-prey behaviour present in many games. Furthermore, it highlights that the CEM weights should not be considered a burden, but rather a feature to create different personas, thus increasing the believability of our NPCs while overcoming the weaknesses of utility-based agents outlined in the introduction. 

\subsection{Experiment II: Exploiting Affordances}
In a sufficiently complex game, the wealth of possible interactions between a character's abilities and features of the environment becomes hard to anticipate even for the game's designers. As a consequence, most hand-crafted NPCs do not fully exploit these interactions. In more open-ended algorithms such as MCTS, this anticipation problem creeps into the definition of the optimisation objective, resulting in blunt adversary behaviour. Empowerment is defined on an agent's possible interactions with its world, and should thus be sensitive to \emph{any} interaction between \emph{any} type of ``functional content'' \cite{smith2014understanding}. In our second experiment, we thus investigate if a CEM-driven NPC can leverage the possible interactions that a game affords to the full extent for adversarial behaviour. Because tiny changes to e.g. the environment or an agent's abilities can turn the emerging gameplay upside down, we start with a simple environment and extend it gradually to investigate CEM-driven adaptation.

\begin{figure*}[htbp]
\resizebox{\linewidth}{!}{
\centering
\begin{subfigure}[b]{.19\linewidth}
  \includegraphics[width=\linewidth]{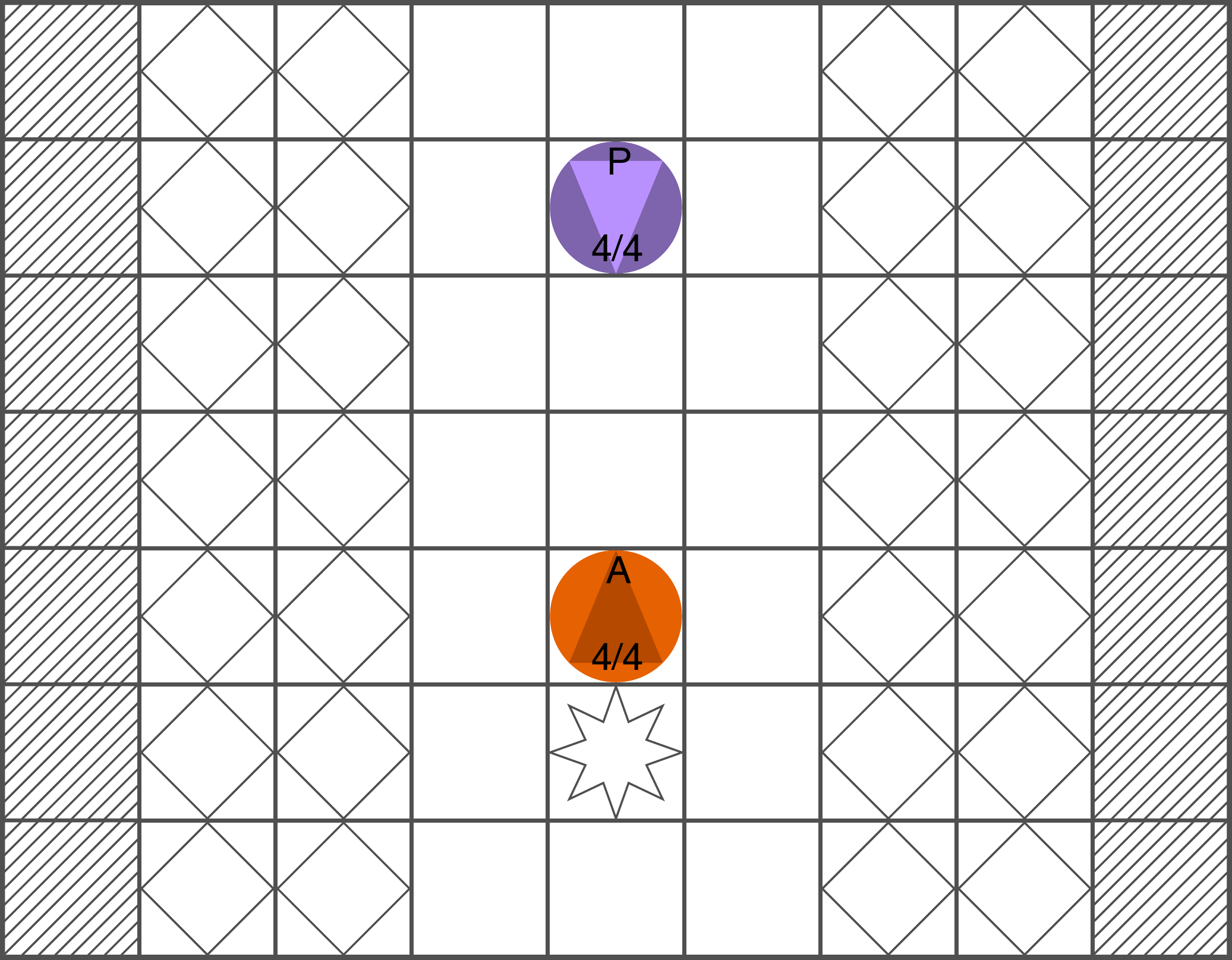}
\end{subfigure}
\begin{subfigure}[b]{.19\linewidth}
  \includegraphics[width=\linewidth]{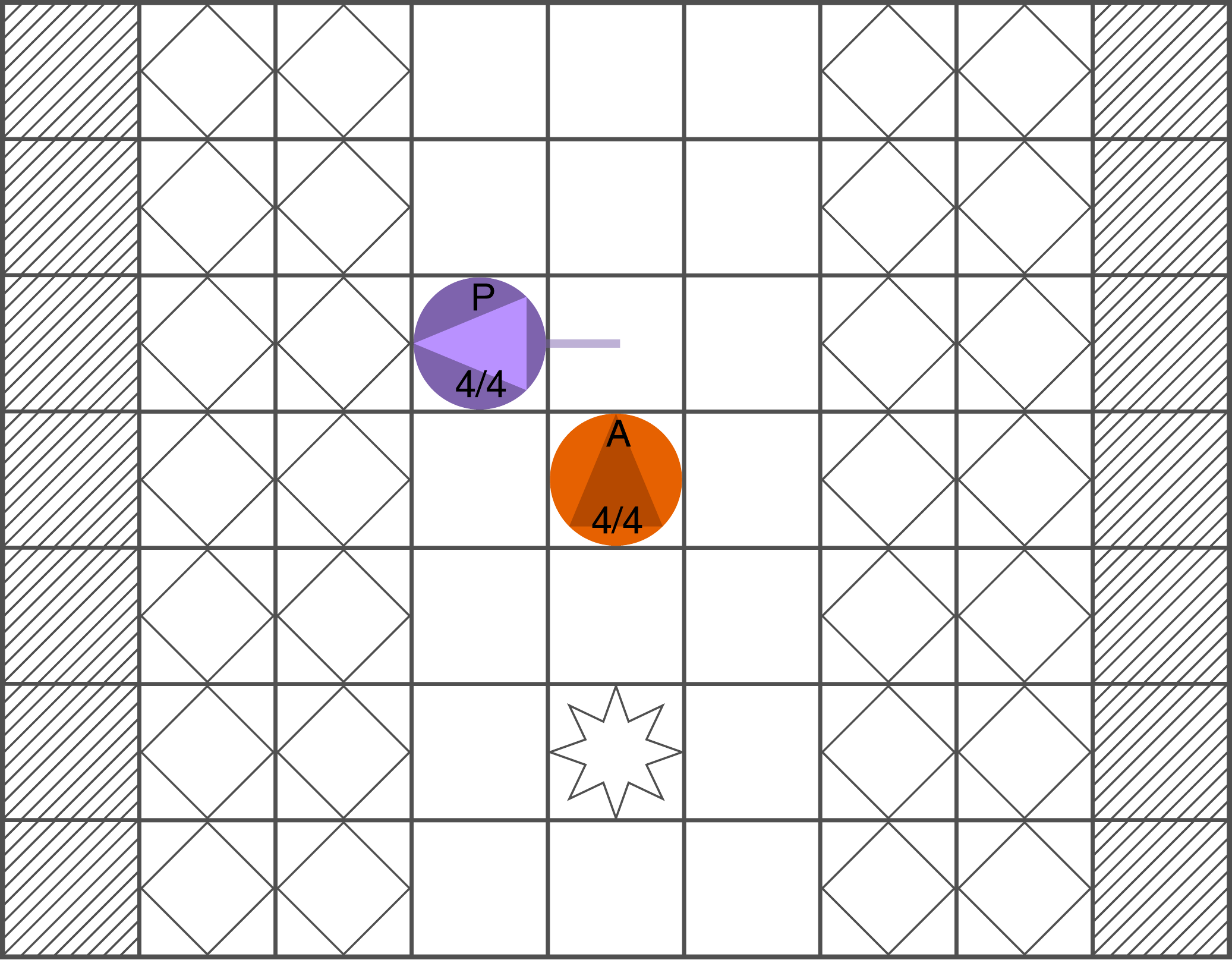}
\end{subfigure}
\begin{subfigure}[b]{.19\linewidth}
  \includegraphics[width=\linewidth]{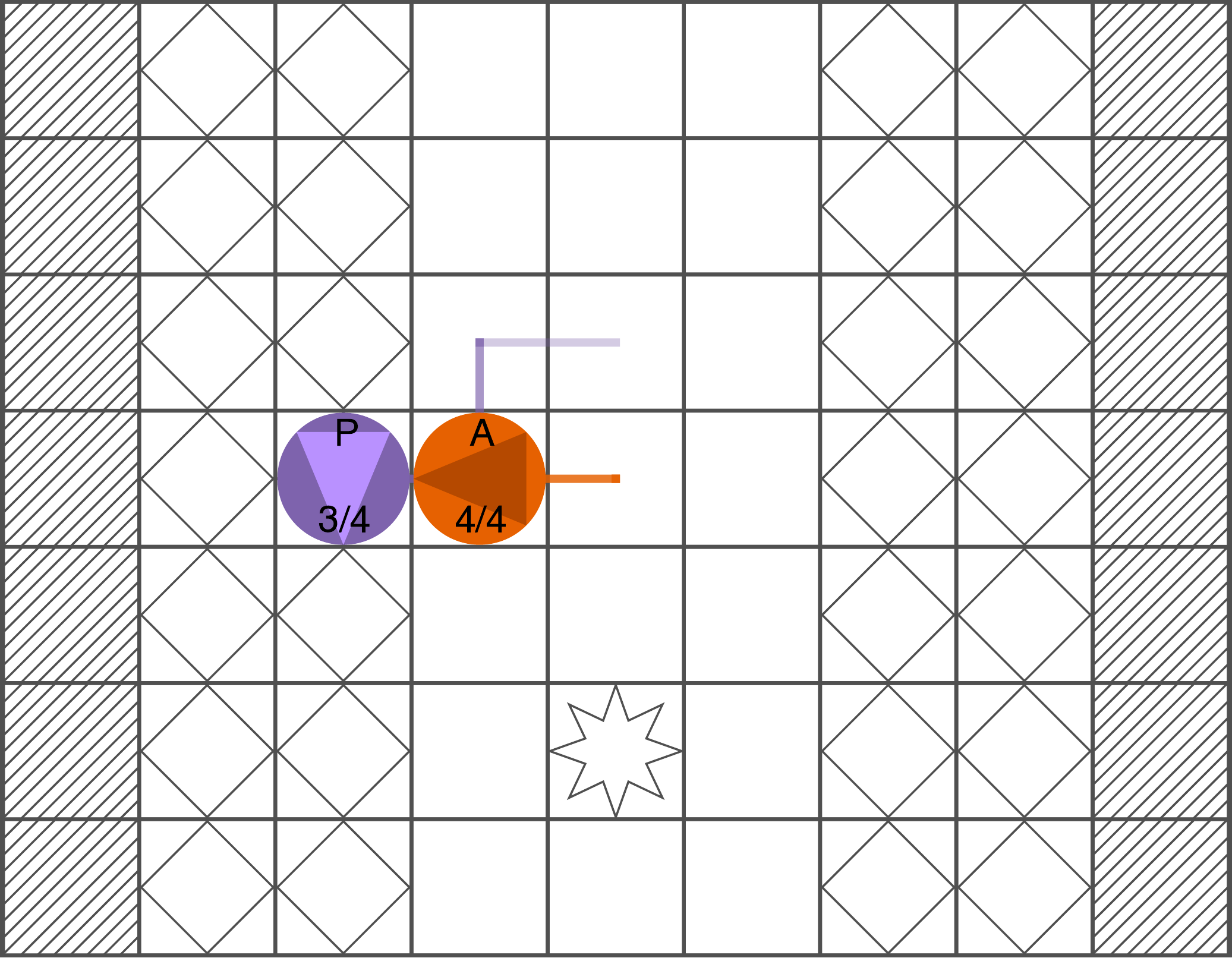}
\end{subfigure}
\begin{subfigure}[b]{.19\linewidth}
  \includegraphics[width=\linewidth]{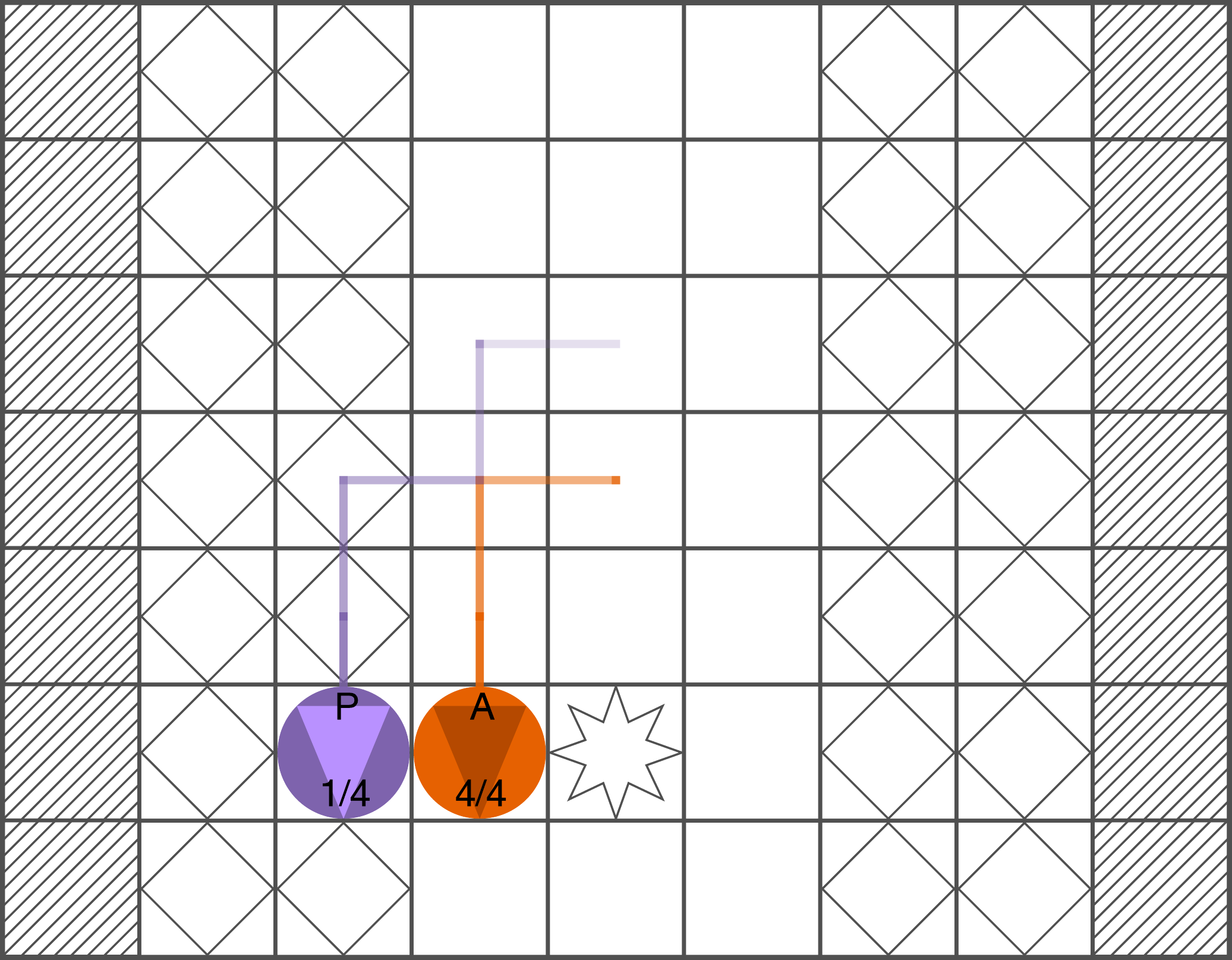}
\end{subfigure}
\begin{subfigure}[b]{.19\linewidth}
  \includegraphics[width=\linewidth]{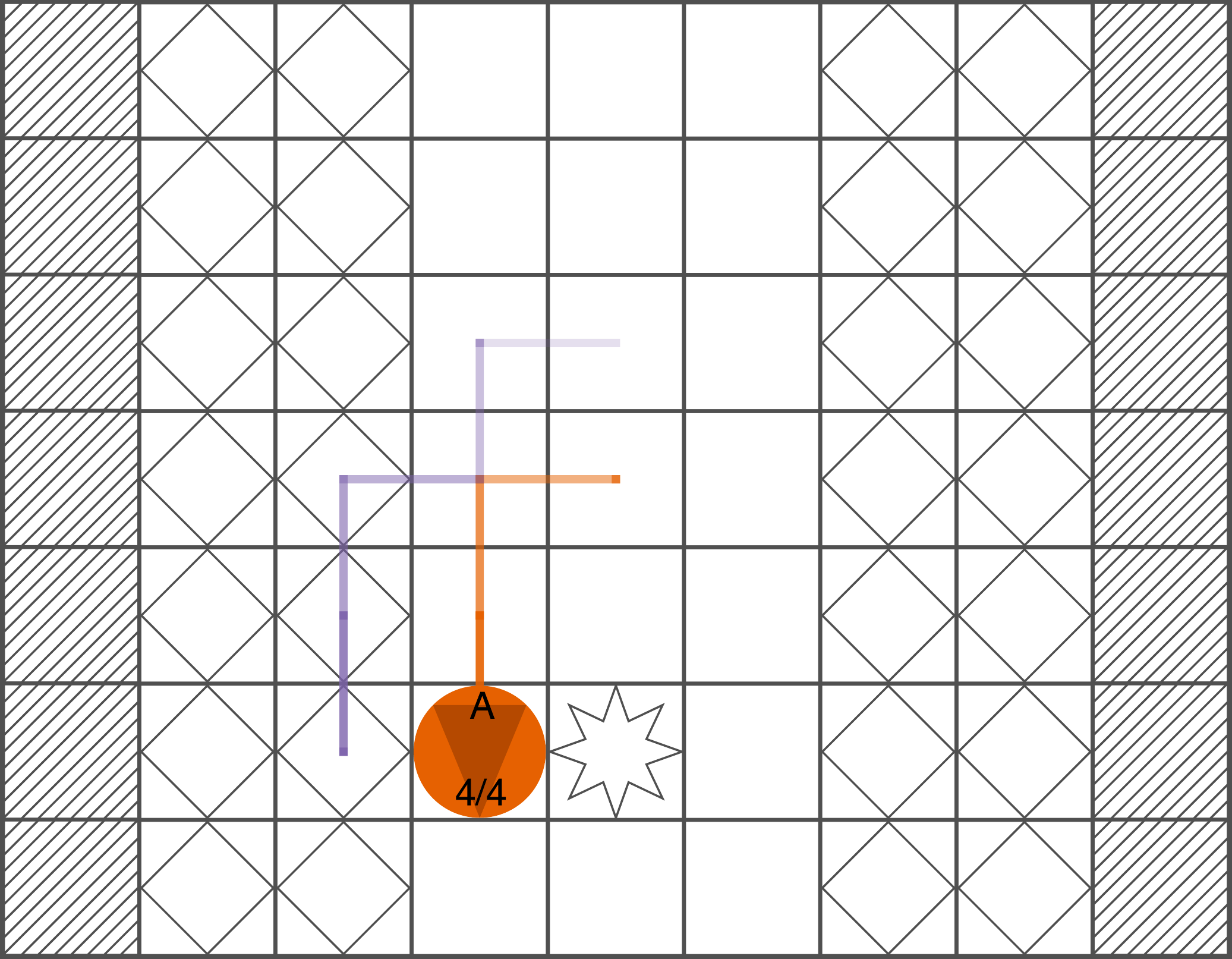}
\end{subfigure}
}
\caption{Experiment 2 (Detail). Adversary pushing player into lava and blocking it from returning to the platform.}
\label{fig:exp_exploiting_push_into_lava}
\vspace{-0.3cm}
\end{figure*}

\begin{figure}[htbp]
\centering
\begin{subfigure}[b]{.4\linewidth}
  \includegraphics[width=\linewidth]{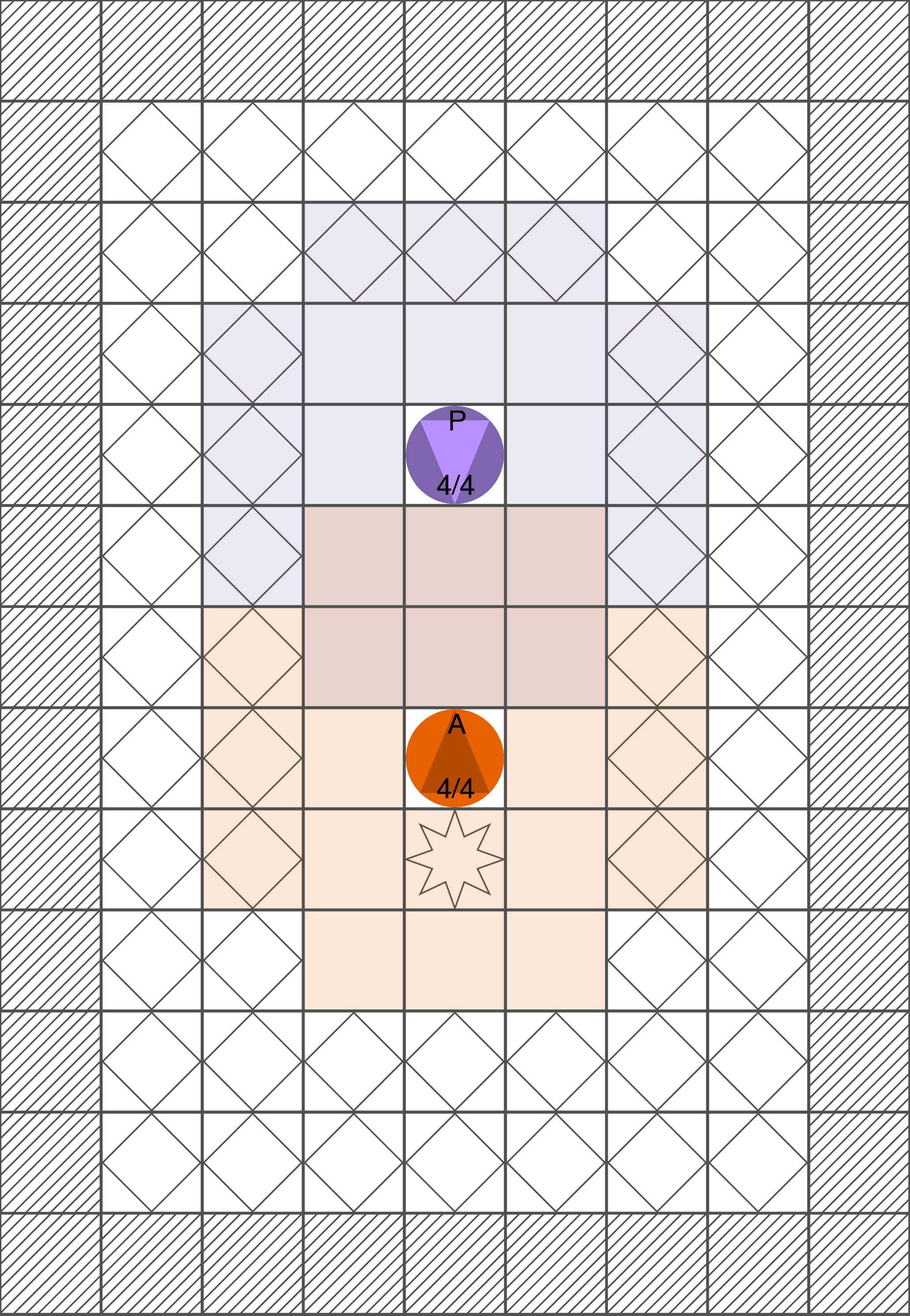}\vspace*{0.03cm}
  \caption{Initial state}
  \label{fig:exp2Initial}
\end{subfigure}
\begin{subfigure}[b]{.4\linewidth}
  \includegraphics[width=\linewidth]{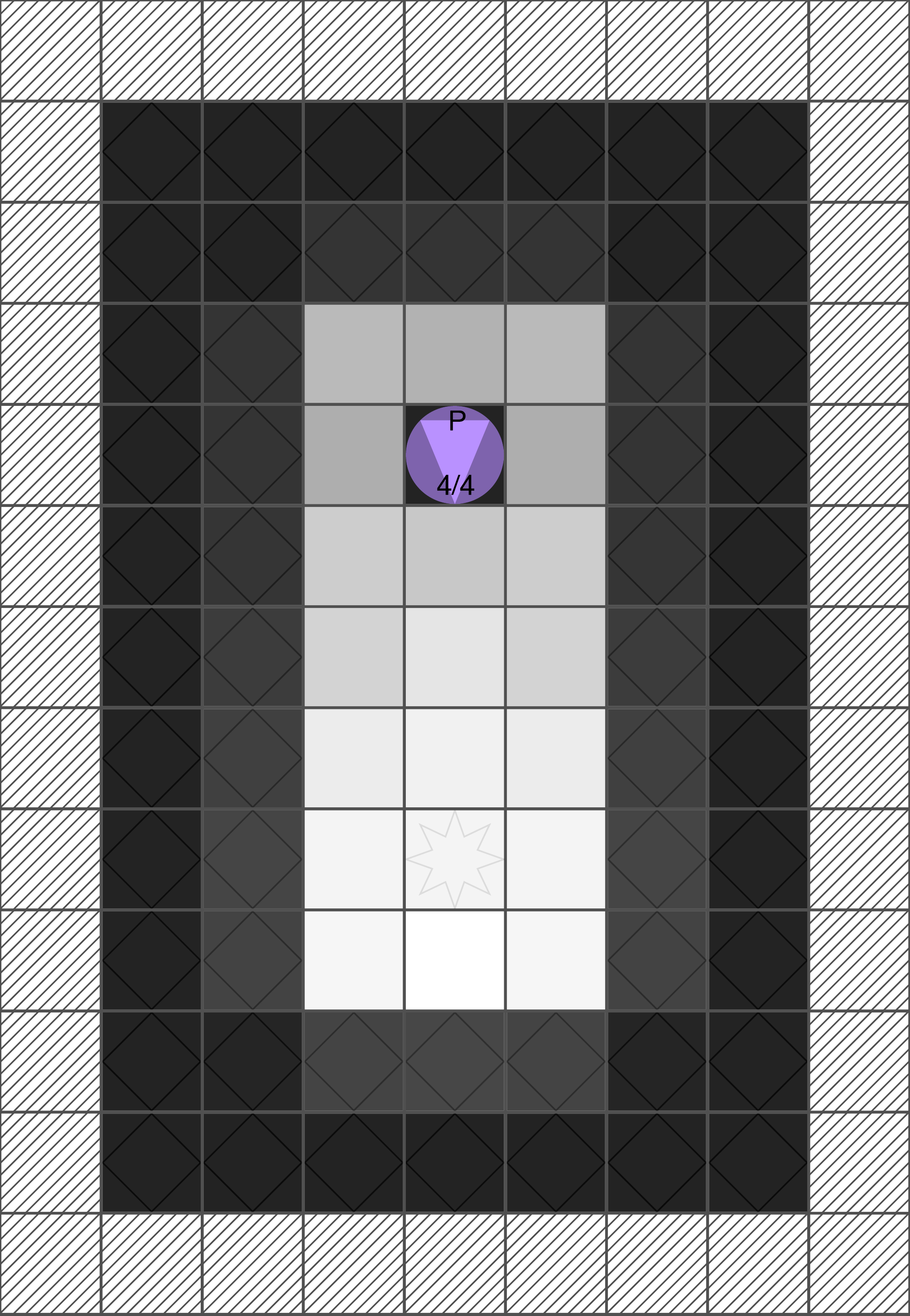}
  \caption{$\mathfrak{E}^{A,3}$}
  \label{fig:exp2e3step}
\end{subfigure}\\[2ex]
%\hfill
\begin{subfigure}[b]{.4\linewidth}
  \includegraphics[width=\linewidth]{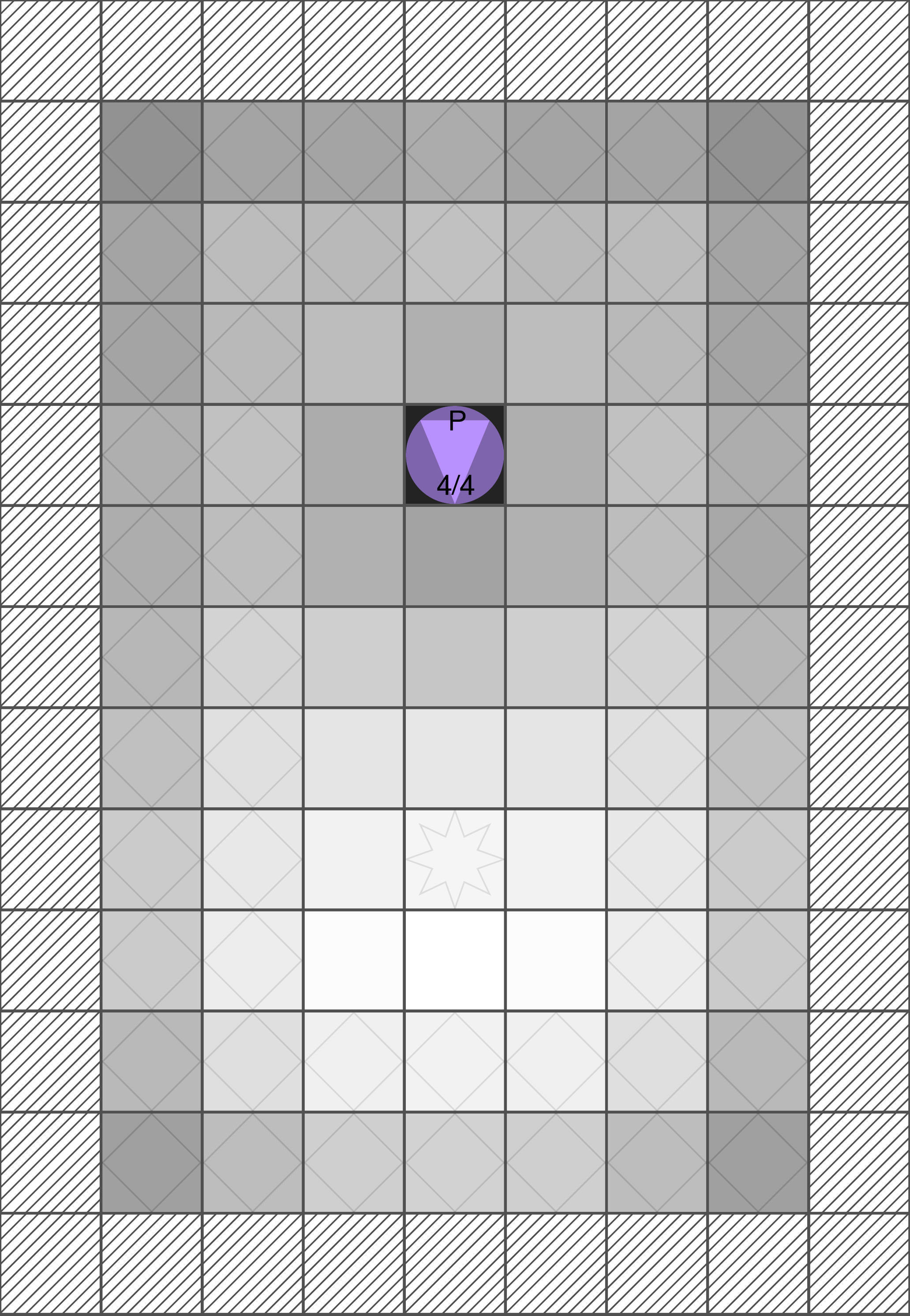}
  \caption{$\mathfrak{E}^{A,3}$, flying}
  \label{fig:exp2e3stepFlying}
\end{subfigure}
\begin{subfigure}[b]{.4\linewidth}
  \includegraphics[width=\linewidth]{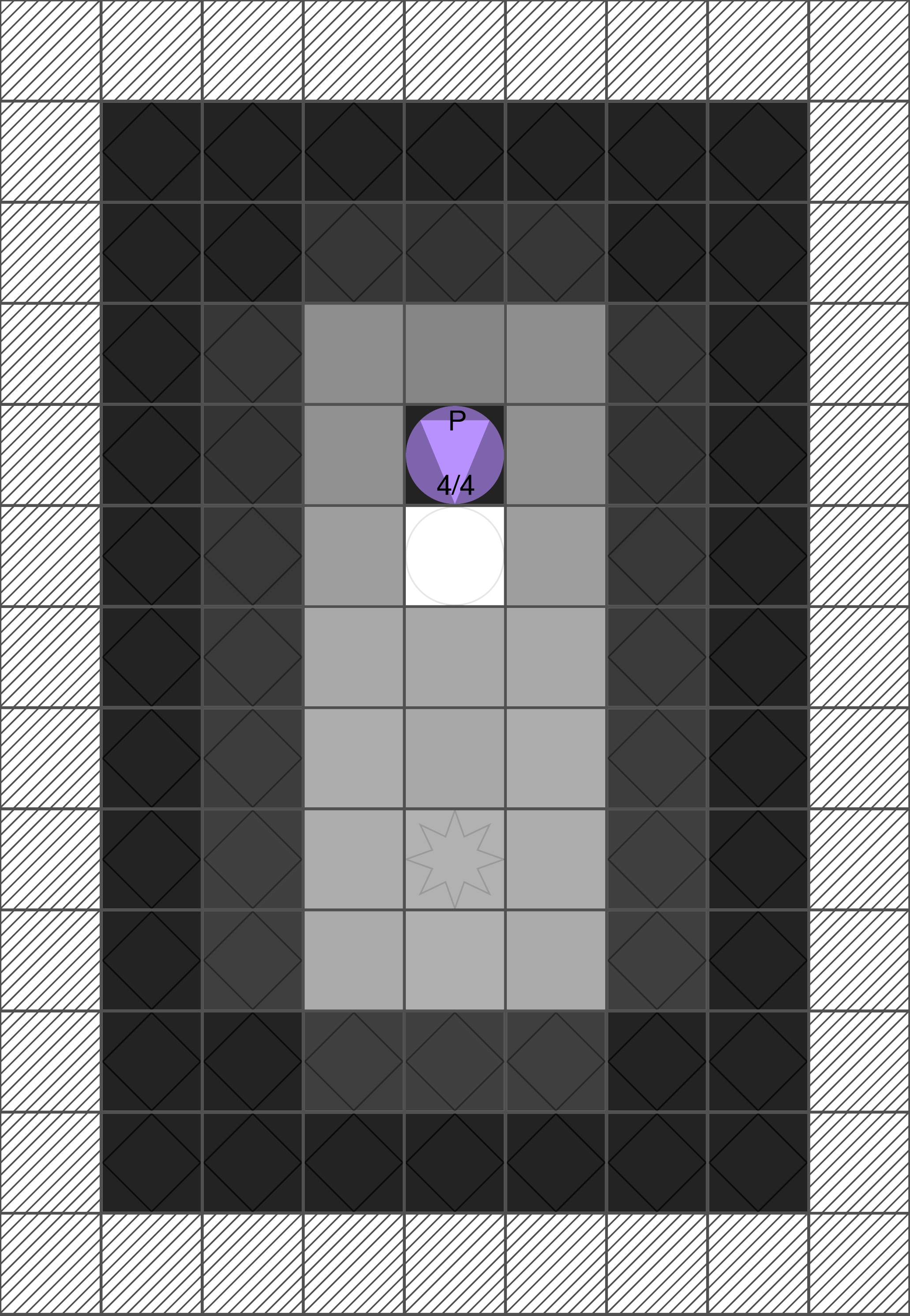}
   \caption{$\mathfrak{E}^{A,3}$, recharger}
   \label{fig:exp2e3stepRecharger}
\end{subfigure}
\caption{Experiment 2. Initial state with perceptive field and 3-step empowerment for different modifications.}
\label{fig:exp_exploiting_env_1}
\vspace{-0.3cm}
\end{figure}

Fig. \ref{fig:exp2Initial} shows the initial state of the environment, where adversary and player face each other in an arena surrounded by lava. If an agent happens to step on the lava, its health decreases by one unit per time step. In order to examine longer interaction sequences, we extend our characters' health to four units ($h_t, h_{\textit{max}}=4$). Mediated by \emph{health-performance-consistency}, a decrease in health results in lower empowerment even for small lookaheads. The NPC's 3-step empowerment (Fig. \ref{fig:exp2e3step}) is thus lower in the lava, and decreases the further away the agent is from the platform, where only few action sequences lead back alive.

Under the default configuration, the NPC closes up to the player and blocks it to reduce the latter's movement and thus empowerment. If we give the NPC the ability to push though, the dynamics change considerably: As illustrated in Fig. \ref{fig:exp_exploiting_push_into_lava} and in a video\footnote{Video online: \url{youtu.be/-Stm59llrDs}}, the NPC then destroys the player by pushing it into the lava. Importantly, it blocks the player from returning to the platform, no matter which path the latter chooses. The policy thus captures how the agent's new ability, in interaction with the environment, can be exploited the decrease the player's empowerment -- resulting in more challenging and arguably novel gameplay.

But what happens if we give the NPC an action which is typically not associated with adversaries, such as healing? Our next scenario shows that this surprisingly takes the adverseness of our NPC to a new level: equipped with the ability to heal the player by one health unit per time step, it still pushes the player into the lava. However, once the player is close to ceasing, the NPC uses its healing action to keep the player alive\footnote{Video online: \url{youtu.be/fy-2hRf-4L8}}. Crucially, the player's health in this situation would be too low to make it back to the platform. Our CEM-driven NPC thus acts in best super-villain style, and in stark contrast to e.g. MCTS with the only objective to destroy: it just keeps the player's health sufficiently high to exercise control over it -- thus optimising its own empowerment -- while keeping the player's empowerment low. We can modulate this behaviour by changing the weight parameters: if we reduce $\alpha_A$, the NPC lets the player die.

\begin{figure*}[htbp]
\resizebox{1.003\linewidth}{!}{\hspace{-0.08cm}
\centering
\begin{subfigure}[b]{.199\linewidth}
  \includegraphics[width=\linewidth]{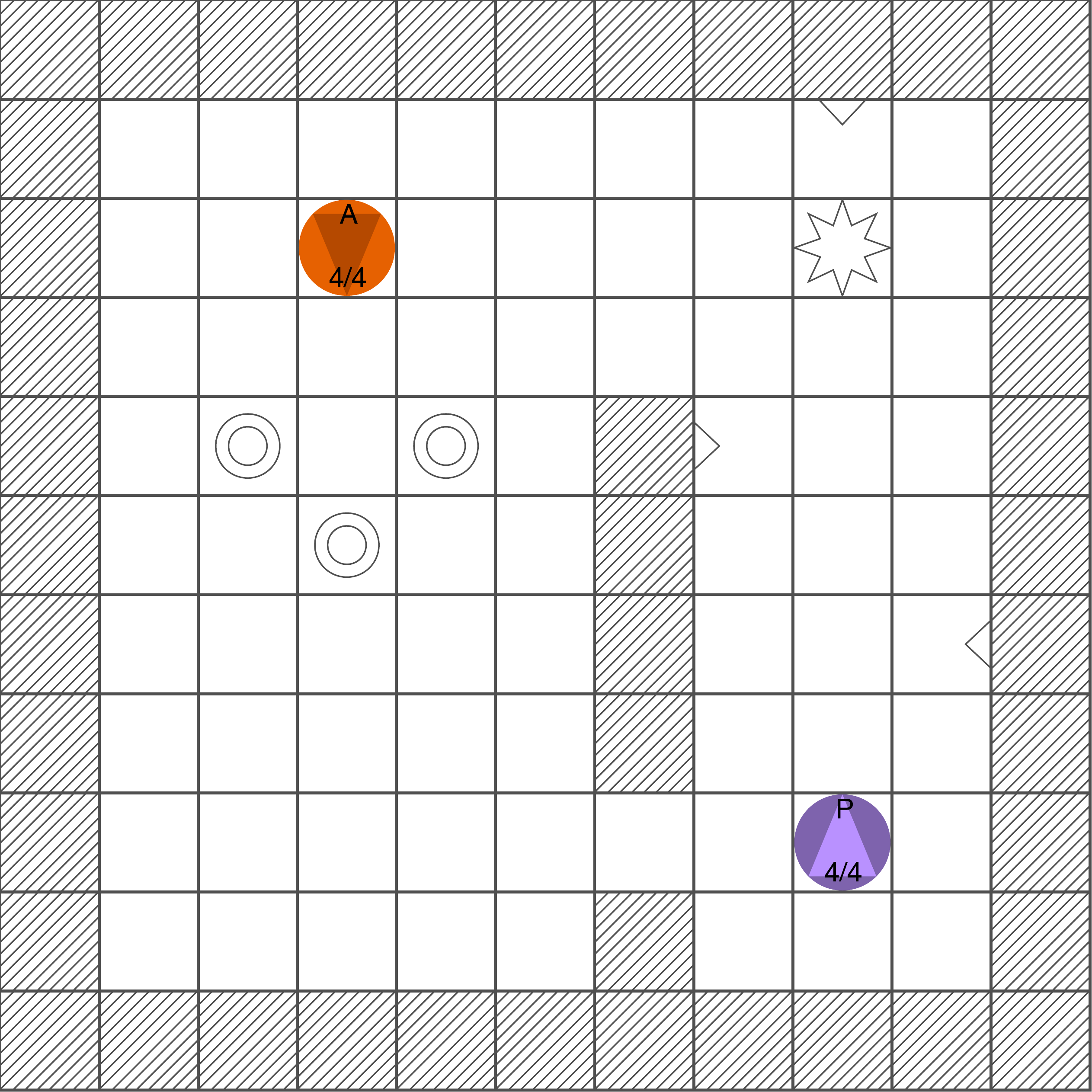}
\end{subfigure}
\begin{subfigure}[b]{.199\linewidth}
  \includegraphics[width=\linewidth]{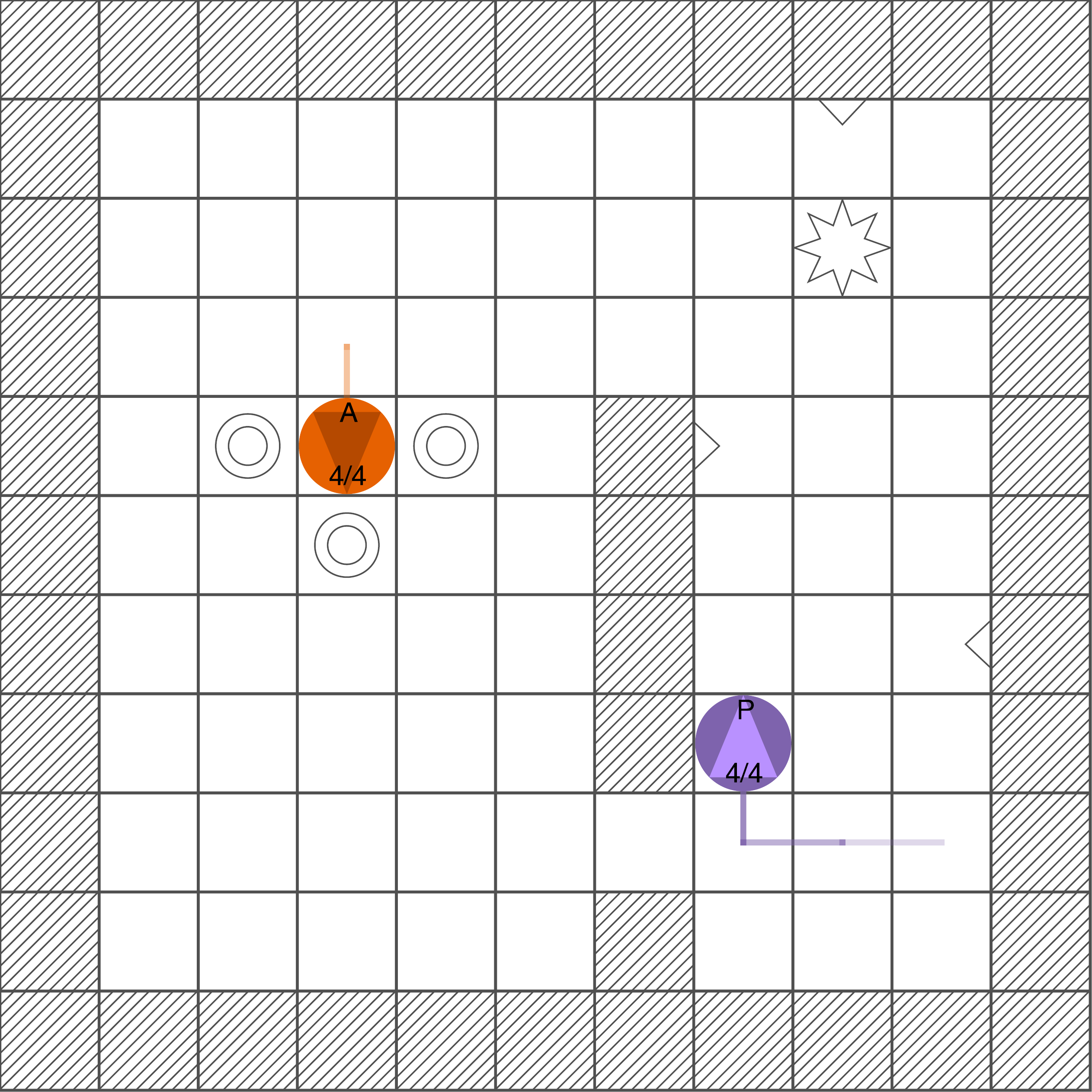}
\end{subfigure}
\begin{subfigure}[b]{.199\linewidth}
  \includegraphics[width=\linewidth]{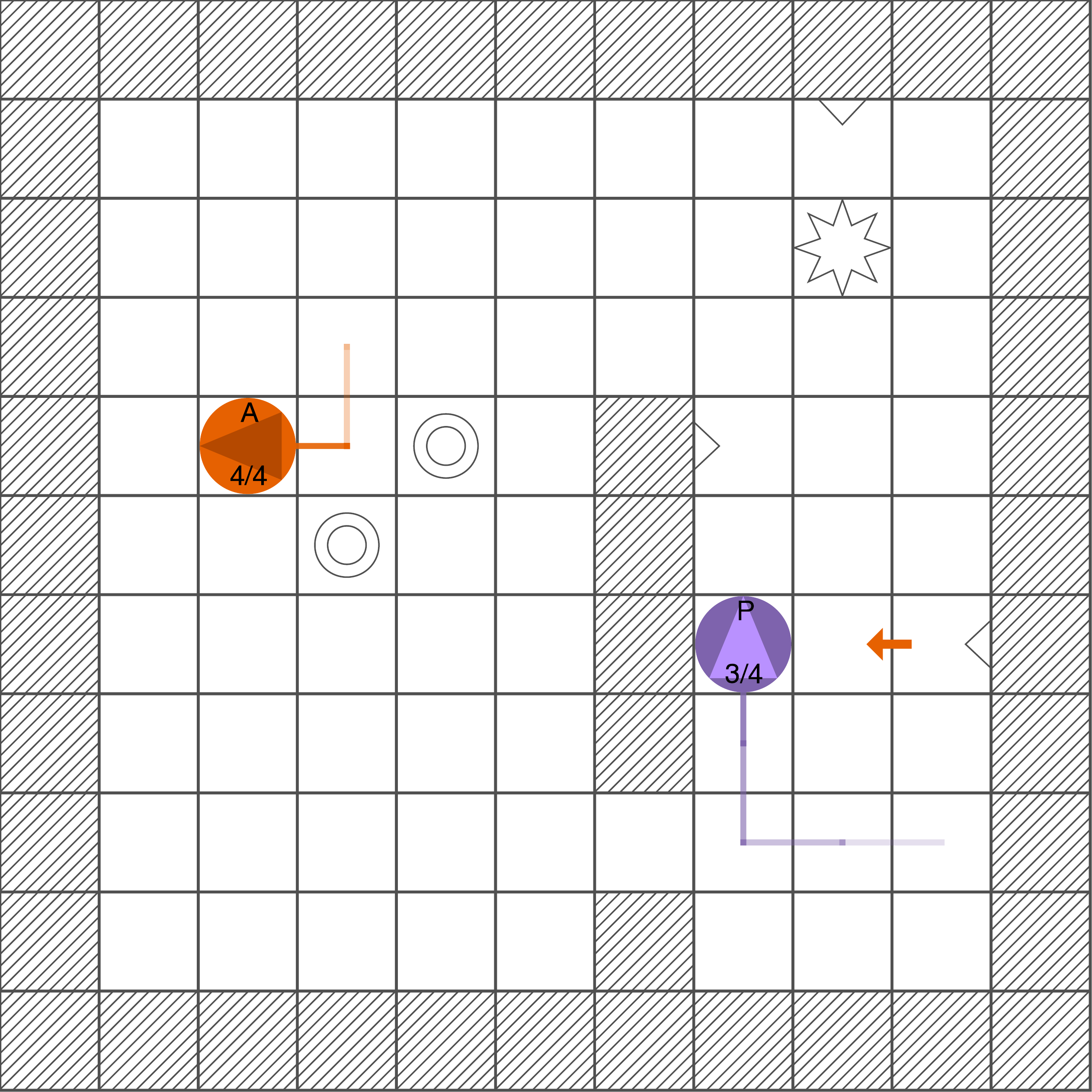}
\end{subfigure}
\begin{subfigure}[b]{.199\linewidth}
  \includegraphics[width=\linewidth]{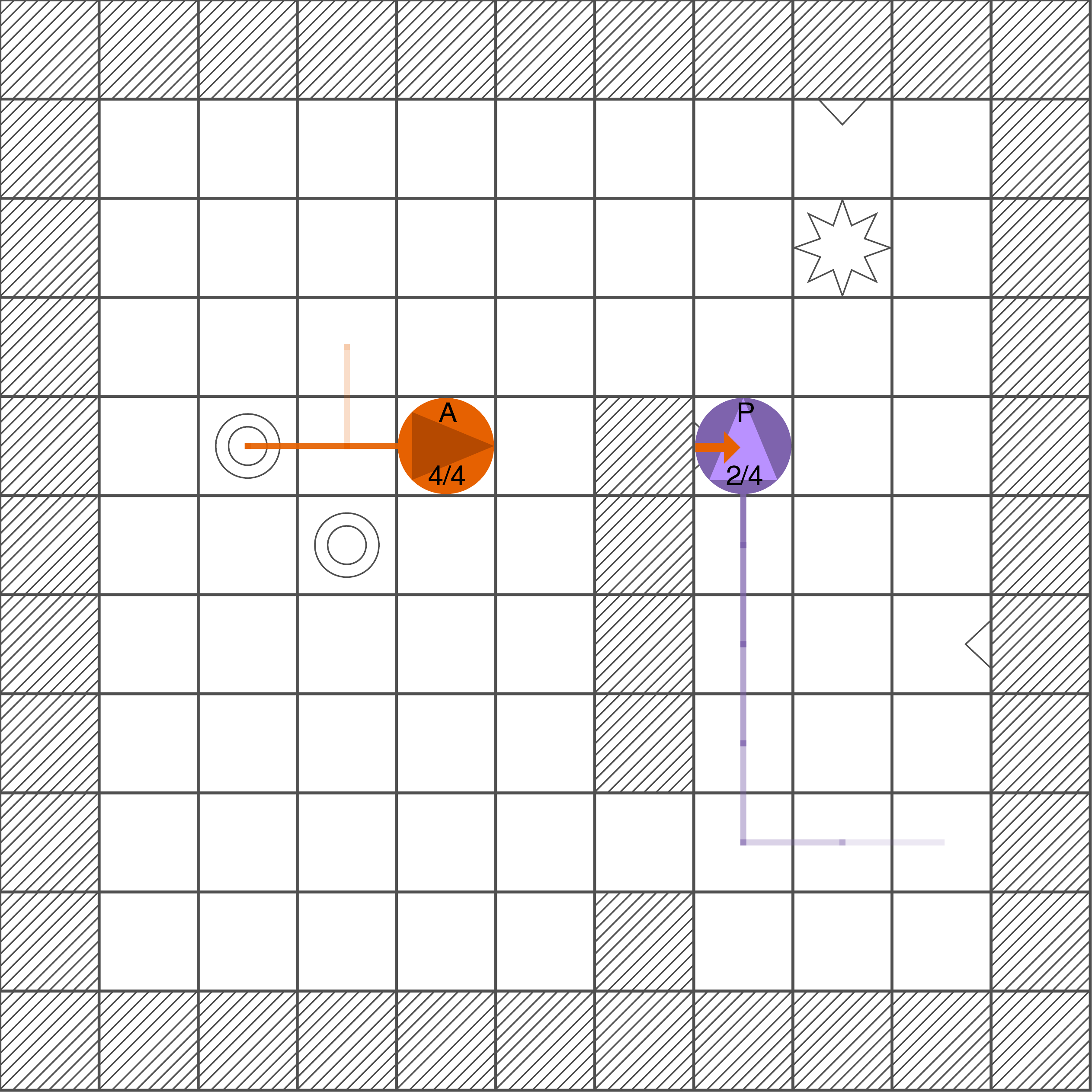}
\end{subfigure}
\begin{subfigure}[b]{.199\linewidth}
  \includegraphics[width=\linewidth]{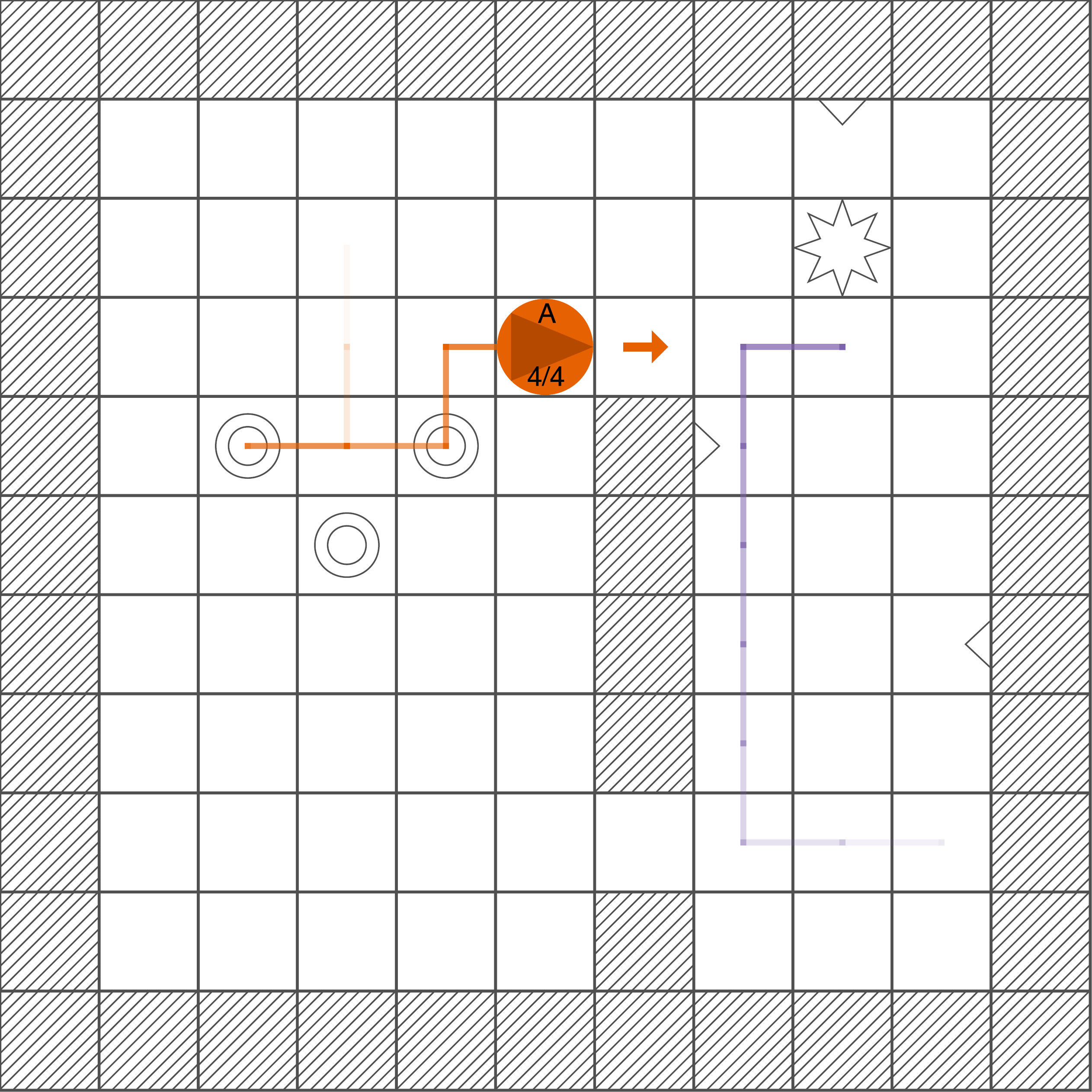}
\end{subfigure}
}
\caption{Experiment 3. Adversary harming player by triggering turrets remotely, eventually destroying it with a range attack.}
\label{fig:exp_distant_threats}
\vspace{-0.3cm}
\end{figure*}

\begin{figure}[htbp]
\centering
\begin{subfigure}[b]{.49\linewidth}
  \includegraphics[width=\linewidth]{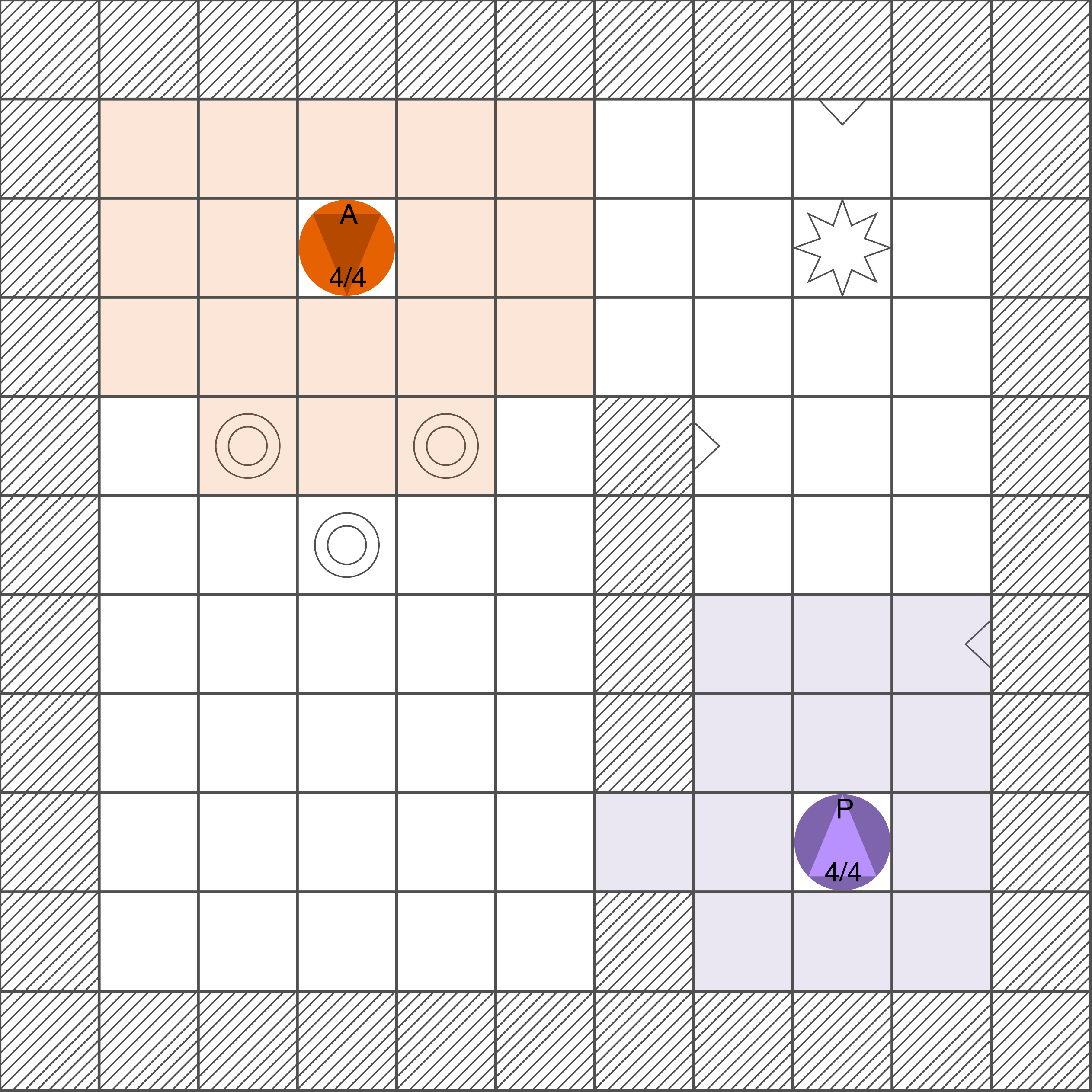}\vspace*{0.03cm}
  \caption{Initial state}\label{fig:experiment_distant_initial}
\end{subfigure}
\begin{subfigure}[b]{.49\linewidth}
  \includegraphics[width=\linewidth]{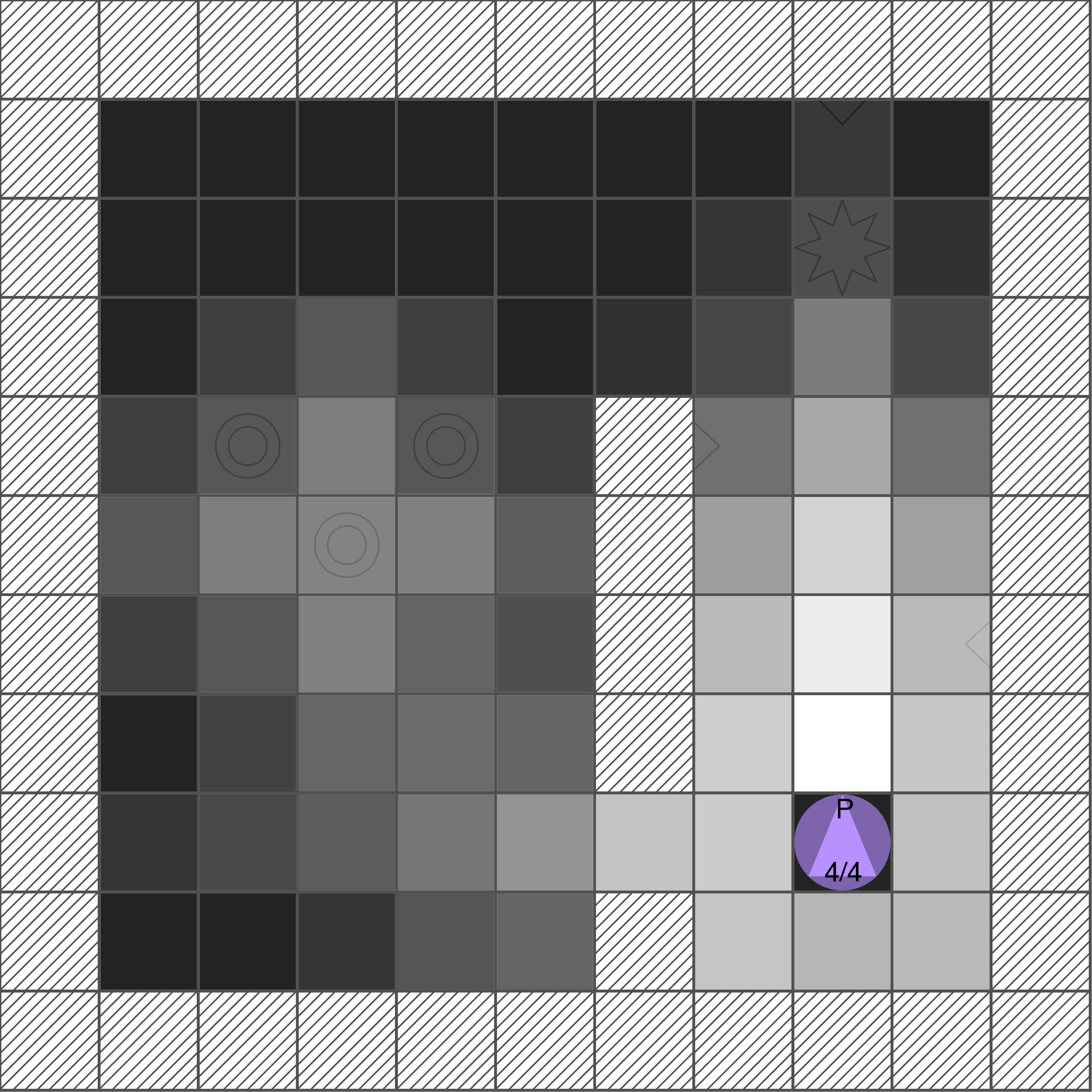}
  \caption{$\mathfrak{E}^{T,3}$}\label{fig:experiment_distant_eT_n3_1}
\end{subfigure}\\[2ex]
\hfill
\begin{subfigure}[b]{.49\linewidth}
  \includegraphics[width=\linewidth]{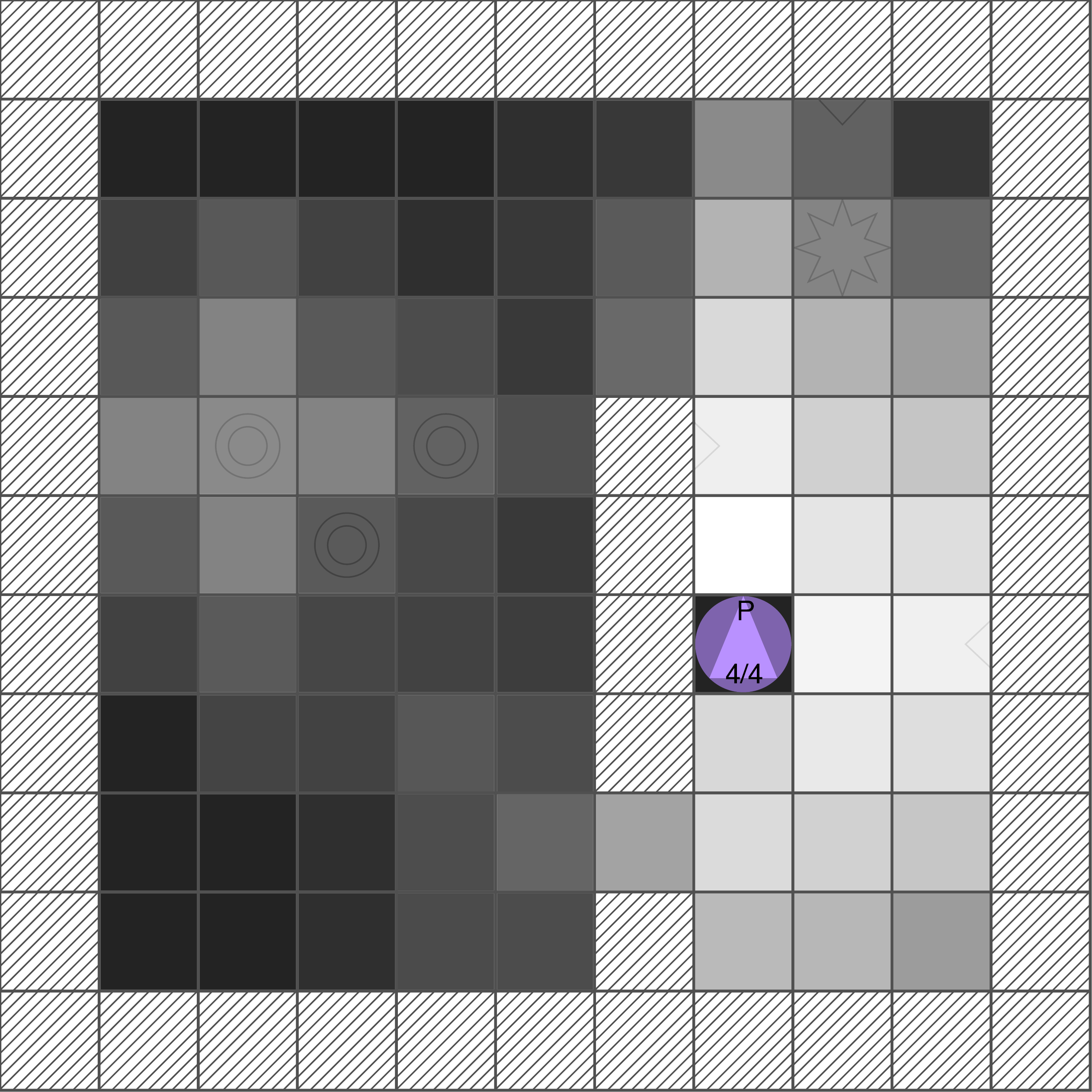}
  \caption{$\mathfrak{E}^{T,3}$}\label{fig:experiment_distant_eT_n3_2}
\end{subfigure}
\begin{subfigure}[b]{.49\linewidth}
  \includegraphics[width=\linewidth]{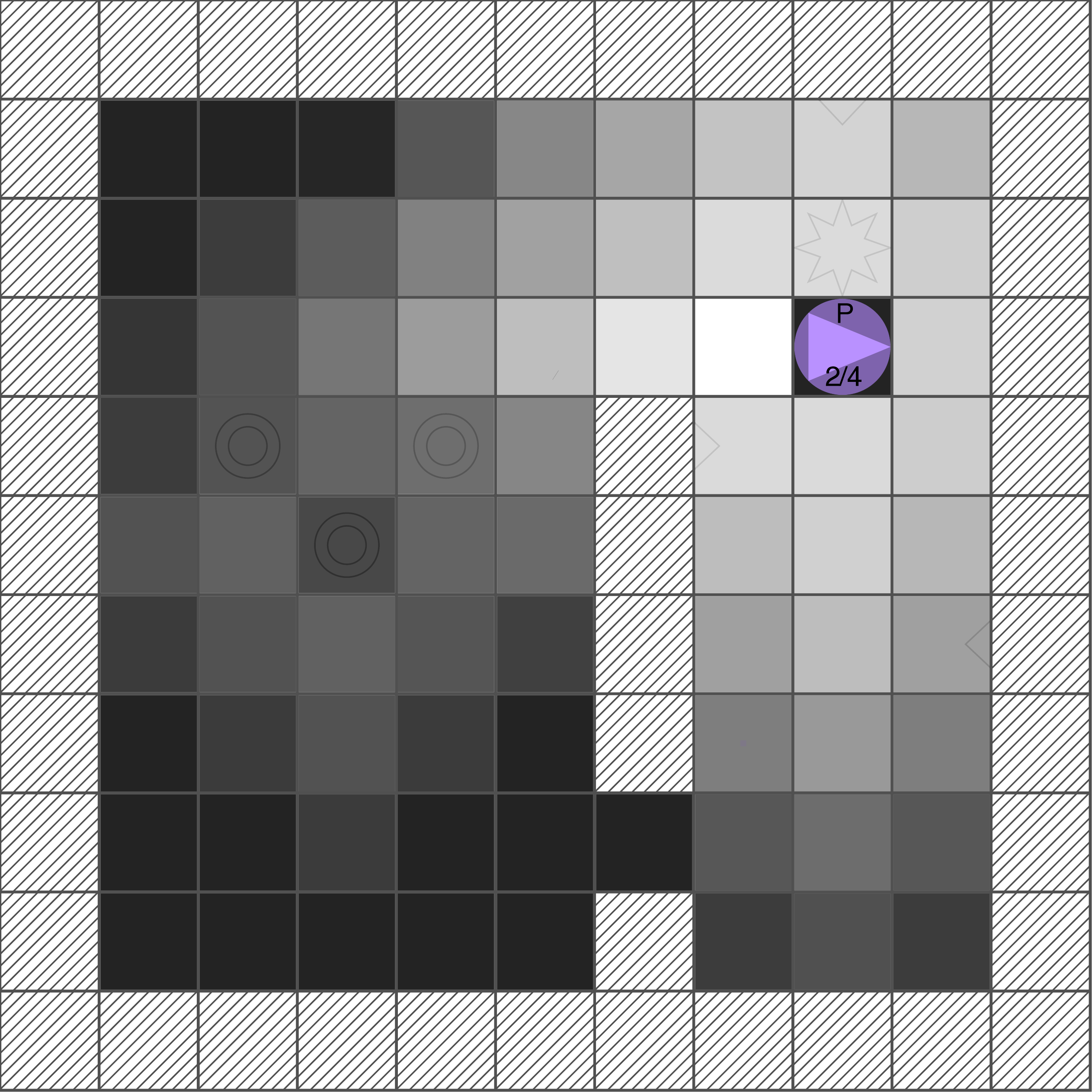}
  \caption{$\mathfrak{E}^{T,3}$}\label{fig:experiment_distant_eT_n3_3}
\end{subfigure}
\caption{Experiment 3. Initial state, followed by adversary $\mathfrak{E}^{A,n}$ and adversary-player transfer empowerment $\mathfrak{E}^{T,n}$, the latter for three different player positions. Lookahead $n=3$.}
\label{fig:exp_distant_threats_maps}
\vspace{-0.5cm}
\end{figure}

Maximising empowerment cannot only be achieved by controlling other characters; in scenarios like the present, it also requires to engage in acts of self-preservation. Dodging attacks by the player as in the previous experiment is such an act. However, previously both agents had identical abilities, which is uncommon in most games. To examine whether CEM can exploit inequalities between characters to further both self-preservation and adverseness, we allow our NPC to range-attack and fly, while the player is limited to melee attacks on the ground. In our testbed, a character that can fly is not affected by the hazardous effect of lava, and the NPC's empowerment is thus not affected by the lava anymore, but only by the surrounding walls and the player (cf. Fig. \ref{fig:exp2e3step} and \ref{fig:exp2e3stepFlying}). With its new ability, our NPC now dodges the player's melee attacks by escaping over the lava. Once the player veers away from the NPC, it returns and attacks from a distance\footnote{Video online: \url{youtu.be/tSzYLaCDXiI}}. Note that, using a uniform model of the player's policy, the NPC expects the player to remain on the platform no more than following it into the lava. However, this would result in a decrease of the player's empowerment -- which would be welcome to the NPC trying to minimise it.

In our last scenario, we stress another aspect of self-preservation: not escaping harm, but recovering from it. If we allow our characters to push and perform melee attacks, the NPC engages in close combat, using both direct attacks and pushing the player into the lava. Meanwhile, if we put a recharge unit in the middle of the platform, the characters start competing for the scarce resource: once the NPC's health gets close to zero, it captures the recharge tile to recover, pushing the player off if necessary\footnote{Video online: \url{youtu.be/WoWfLRlY2LY}}. The NPC's empowerment heatmap for a lower health state $h_t=2, h_{\textit{max}}=4$ (Fig. \ref{fig:exp2e3stepRecharger}) highlights the effect of the recharge station - like a beacon in the reward landscape. This second experiment support our hypothesis that CEM-driven agents can adapt to- and exploit changes in the environment and in their abilities.

\subsection{Experiment III: Distant Threats}
The most challenging adversaries are arguably those that strike from a distance, where they remain unaffected by our actions, and potentially also undetected. An NPC that would be inferior in direct combat could cast spells, order air strikes or control traps and doors remotely. In our last experiment, we investigate if CEM allows for such behaviour to emerge.

Key to such behaviour is player and transfer empowerment, with transitions towards direct interactions being facilitated by trading off the NPC's own empowerment. We have designed our last experiment to provoke such a transition and examine the interplay of these three components. In the initial state (Fig. \ref{fig:experiment_distant_initial}), the player starts on the lower right in a corridor, while the NPC is situated on the upper left in an open area, separated by a wall with two passages. The player faces three turrets, two on the sides and one ahead. The corresponding triggers are positioned in front of the NPC. Both characters have the ability to perform a range attack.

The NPC's own empowerment in this state does not convey any information about the best trigger to affect the player, as it only quantifies the NPC's influence on its \emph{own} sensor state. Player- and transfer empowerment in contrast both work as proxy to the player's condition: transfer empowerment measures the impact of turret-triggering on the player's health, which is captured in the latter's sensor; the player's health in turn affects the player's empowerment, which can be exploited by the NPC. Figs. \ref{fig:experiment_distant_eT_n3_1} -- \ref{fig:experiment_distant_eT_n3_3} show how transfer empowerment peaks on- and around the triggers for player positions in the shooting range of different turrets. 

Following the CEM policy, the NPC triggers the correct turrets to hit the player on its way towards the goal tile (Fig. \ref{fig:exp_distant_threats} 1--4). When the player moves between turrets, the NPC positions itself where it can strike quickest, i.e. between the triggers. Once the player gets closer to the goal and thus to the open passage towards the adversary, the latter trades off its own- against the increase of transfer empowerment: the spatial proximity of the player results in a transfer empowerment gradient which the NPC could follow to eventually attack the player directly. By doing so however, the NPC risks its own empowerment to be decreased by a counter-attack. In the present configuration, the adversary eventually moves away from the triggers and attacks the player directly (Fig. \ref{fig:exp_distant_threats}, last and video\footnote{Video online: \url{youtu.be/qBTdGCkspA4}}). Meanwhile, decreasing the NPC's health ($h_t{=}1, h_{\textit{max}}{=}2$) makes it remain at its current position and shoot the player from a distance\footnote{Video online: \url{youtu.be/HnVE-IHmGG8}}. This experiment supports that CEM also yields complex remote interactions.

\section{Discussion}
Our CEM-driven NPC not only proved to be very sensitive to changes in the environment and its own abilities; we also demonstrated how small modifications of the weights can switch behavioural patterns, and yield different character types such as opportunists, daredevils and ``super-villains''. 

Our experiments however have also pointed out the importance of incorporating stronger assumptions about the player's policy to yield more believable behaviour. At present, the NPC assumes all player actions in a given state to be equally likely. Thus, while the CEM policy equips the NPC with a drive for survival and self-defense, no such assumption is present in the model of the player's policy. More than that, the adversary-player relationship is one-sided: while the NPC would select its actions to diminish the player's empowerment, the player is not assumed to have a negative bias. We think that empowerment can be used successfully to induce such a bias into the NPC's model of the player's policy, while maintaining the generality of the approach. Importantly, assuming the player to minimise the NPC's empowerment would be short-sighted: unless fighting adversaries contributes explicitly to achieving a game's goal, a human player might be more inclined to evade adversaries than to attack them. Instead, we suggest to go one step further and model the player as maximising empowerment itself. This should yield a good prior particularly in games where progress is aligned with an increase in options and influence. We presently do not represent the player's goal in the policy model, but CEM operates implicitly on the player's trajectories towards goal achievement. Ultimately, we expect the quality of adversarial behaviour to increase further when inferring and modelling these goals explicitly.

Another question arising from our experiments is how challenge induced by CEM-driven adversary NPCs can be modulated to produce a well-balanced player experience \cite{denisova2017challenge}. Given that our NPCs adapt to changes in their abilities, such modulation can be facilitated by the classic means of balancing the characters' abilities with respect to the environment. But CEM offers additional alternatives. Adjusting the weight parameters allows us to model characters that challenge us in different ways: an adversary can be made aggressive or more cautious, only fighting back if they are confronted directly. Furthermore, noise can be introduced into the NPCs model of the environment dynamics, making it overconfident or insecure about their own and other characters' possible interactions with the world. Finally, biases can also be introduced into the NPC's model of the player's policy, rendering the latter e.g. as anticipated threat or harmless peer. We expect this to yield particularly interesting gameplay in combination with online model learning.

We finally want to address the scalability of CEM. In order to discriminate small effects of the underlying quantities in this study, we have computed coupled empowerment exhaustively. However, this comes with exponential computational complexity, mostly due to the calculation of the forward transitions and the channel capacity. In recent years, several approximations for the maximisation of mutual information, underlying empowerment, have been proposed, drawing on variational inference and deep neural networks  \cite{Shakir2015, gregor2016variational, karl2017unsupervised}. We believe that these are presently the most promising candidates to increase the scalability of CEM. Furthermore, the lookahead in CEM can be increased by utilising macro-actions. In sufficiently large action spaces, Monte-Carlo sampling of action sequences (cf. \cite{salge2014changing}) is also likely to yield good approximations. Finally, more informed policy models could not only increase the quality of behaviour, but also be used to prune the search tree.

\section{Conclusion and Future Work}
We have set out to provide an open-ended action policy for NPCs to \emph{leverage any interaction} a game affords, and to \emph{adapt to changes} in a game with the ultimate goal to design more believable characters. In previous work, we have proposed to use CEM to engineer general companion NPCs that yield a large variety of new and potentially surprising, supportive behaviours. In this paper, we have adopted the action policy to give rise to adversarial behaviour. We have shown by means of a qualitative study that minimising the player's empowerment in a CEM policy yields rich adversarial behaviour, based on our NPC's successful exploitation of interaction affordances, and the adaptation to changes to its own- and the player's abilities, as well as to the environment. Our NPC has used its abilities, e.g.~to heal, in ways that we would find surprising even in respect to human opponents.

Our study has provided valuable insights towards increasing the believability of our NPCs further, which we plan to use in a quantitative evaluation to conclude this proof-of-concept. We will employ AI playing agents to remove any experimenter bias, and investigate the suitability of CEM to drive both adversarial and supportive behaviour based on objective metrics such as goal achievement, as well as subjective measures of player experience. Given the present observations, we are confident that our participants will not be bored with stereotypical adversary behaviour, but encounter genuinely new and surprising ways to be mean.

\section*{Acknowledgments}
\small
CG is funded by EPSRC grant [EP/L015846/1] (IGGI). CS is funded by the EU Horizon 2020 programme / Marie Sklodowska-Curie grant 705643. We thank our reviewers for helpful comments.

\bibliographystyle{IEEEtran}
\bibliography{bibliography}

\end{document}